\documentclass[runningheads]{llncs}

\usepackage{eccv}

\newcommand{\eg}{\textit{e.g.}}

\usepackage{graphicx}
\usepackage{microtype}
\usepackage[accsupp]{axessibility}  % Improves PDF readability for those with disabilities.

\usepackage{hyperref}

\usepackage{orcidlink}

\usepackage{subcaption}
\usepackage{soul}
\usepackage{booktabs}
\usepackage{multirow}
\usepackage{amsmath}
\usepackage{enumitem}
\usepackage{pifont}
\usepackage{xcolor}
\usepackage{graphicx}
\usepackage{array}
\usepackage[table]{xcolor}
\usepackage{kotex}
\usepackage{wrapfig}
\usepackage{caption}
\usepackage{titletoc}
\usepackage{algorithm}
\usepackage{algorithmic}
\usepackage{bm}
\usepackage[dvipsnames]{xcolor}
\newcommand{\mm}[1]{\(#1\)}
\newcommand{\mc}[1]{\mathcal{#1}}
\newcommand{\mb}[1]{\mathbb{#1}}

\newcommand{\proposed}{\textsf{DSTP}}

\definecolor{lightskyblue}{rgb}{0.0, 0.75, 1.0}
\definecolor{orangered}{rgb}{1.0, 0.27, 0.0}
\definecolor{violet}{rgb}{0.54, 0.17, 0.89}

\begin{document}
% \setlength{\abovedisplayskip}{3pt} 
% \setlength{\belowdisplayskip}{3pt}  
% \setlength{\abovedisplayshortskip}{0pt}
% \setlength{\belowdisplayshortskip}{3pt} 
\title{Why and When Visual Token Pruning Fails?\\A Study on Relevant Visual Information Shift\\in MLLMs Decoding}
% Stop Reasoning! When Multimodal LLM with Chain-of-Thought Reasoning Meets Adversarial Image


\author{
    \textbf{Jiwan Kim}$^{1}$,
    \textbf{Kibum Kim}$^{1}$,
    \textbf{Wonjoong Kim}$^{1}$, \\
    \textbf{Byung-Kwan Lee}$^{2}$,
    \textbf{Chanyoung Park}$^{1}$\thanks{\small Corresponding author.}
}

\authorrunning{J.~Kim et al.}

\institute{
    $^{1}$Korea Advanced Institute of Science and Technology (KAIST)\\
    \email{\{kim.jiwan, rlqja1107, wjkim, cy.park\}@kaist.ac.kr}\\[0.7em]
    $^{2}$NVIDIA\\
    \email{byungkwanl@nvidia.com}
}

\maketitle

\begin{abstract}
Recently, visual token pruning has been studied to handle the vast number of visual tokens in Multimodal Large Language Models. 
However, we observe that while existing pruning methods perform reliably on simple visual understanding, they struggle to effectively generalize to complex visual reasoning tasks, a critical gap underexplored in previous studies.
Through a systematic analysis, we identify Relevant Visual Information Shift (RVIS) during decoding as the primary failure driver. 
To address this, we propose \textbf{D}ecoding-stage \textbf{S}hift-aware \textbf{T}oken \textbf{P}runing (\proposed{}), a training-free add-on framework that enables existing pruning methods to align visual tokens with shifting reasoning requirements during the decoding stage.
Extensive experiments demonstrate that \proposed{} significantly mitigates performance degradation of pruning methods in complex reasoning tasks, while consistently yielding performance gains even across visual understanding benchmarks.
Furthermore, \proposed{} demonstrates effectiveness across diverse state-of-the-art architectures, highlighting its generalizability and efficiency with minimal computational overhead.
\keywords{Multimodal LLMs \and Visual Token Pruning \and Visual Reasoning}
\end{abstract}

\section{Introduction} \label{sec:intro}
Multimodal Large Language Models (MLLMs)~\cite{llava, qwenvl, internvl} have demonstrated strong capabilities in both \textbf{visual understanding} (e.g., visual question answering (VQA)) and \textbf{visual reasoning}, including visual-centric math, logical puzzles, and STEM-related tasks.\footnote{\scriptsize STEM stands for Science, Technology, Engineering and Mathematics.} 
However, these models typically rely on a massive number of visual tokens generated by a vision encoder~\cite{CLIP, SigLip}.
Such a large number of visual tokens incur prohibitive computational and memory overhead during LLM processing, posing considerable challenges for the practical and efficient deployment of MLLMs.

To mitigate this, a surge of studies~\cite{FastV, SparseVLM, PDrop, VisionTrim} has focused on visual token pruning, aiming to reduce computational overhead by removing redundant visual tokens.
The core objective of these studies lies in detecting visual tokens that contain \textbf{Relevant Visual Information}, which is defined as the essential visual cues required to resolve a given task~\cite{SparseVLM}.
Existing pruning methods typically determine the most relevant visual tokens by leveraging internal MLLM attention~\cite{FastV, PDrop}, vision encoder outputs~\cite{DivPrune, VisionZip}, or a hybrid of both~\cite{SparseVLM, VisionTrim}.
Consequently, by retaining a small number of relevant tokens, these methods significantly accelerate inference while maintaining reliable performance on general visual tasks.
\begin{figure*}[t]
    \centering
    \includegraphics[width=0.98\linewidth]{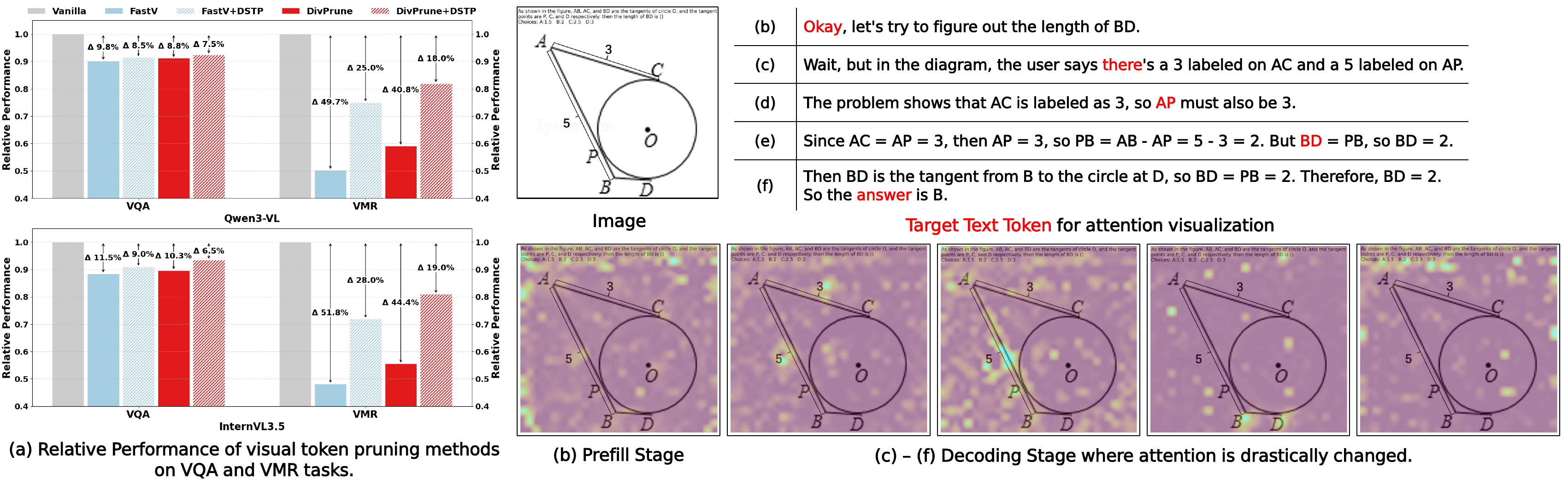}
    \caption{\footnotesize {(a) Performance retention rates of various token pruning methods on Qwen3-VL~\cite{qwen3vl} and InternVL3.5~\cite{internvl3_5} across VQA and VMR.} (b)–(f) Attention heatmaps for a MathVerse~\cite{mathverse} sample on Qwen3-VL relative to the \textcolor{red}{text token} whose attention shifts drastically, reflecting the reasoning context.\protect\footnotemark Full sentences are provided to offer complete context for each reasoning step.}
    \label{fig:intro}
\end{figure*}

However, most existing pruning methods~\cite{FastV, DivPrune, SparseVLM} are primarily evaluated on straightforward visual understanding such as recognizing objects and their attributes using simple VQA benchmarks~\cite{GQA, TextVQA, ScienceQA}.
% However, most existing pruning methods~\cite{FastV, DivPrune, SparseVLM} are primarily evaluated on simple visual understanding benchmarks~\cite{GQA, TextVQA, ScienceQA}, such as recognizing objects and their attributes. 
Under this paradigm,
\footnotetext{\label{note:text_token}\scriptsize Regarding how the specific \textcolor{red}{text token} for the attention heatmap is selected, please refer to~\cref{sec:analysis}.}existing methods have implicitly relied on the expectation that relevant visual information identified during the prefill stage—the initial phase where the model processes the input sequence—would remain sufficient to resolve the given task. 
This assumption holds for simple \textbf{visual understanding} where necessary visual cues are concentrated in narrow regions.
However, as the MLLMs evolve toward reasoning-centric models~\cite{Kimi-VL, GLM4_5V, qwen3vl}, there arises a growing need to address complex \textbf{visual reasoning}. Yet, the efficacy of pruning methods remains largely underexplored in these sophisticated step-by-step reasoning tasks, necessitating the exploration of broader visual regions.\footnote{\scriptsize A detailed comparison between \textbf{visual understanding} and \textbf{visual reasoning} is in~\cref{app:A_task_diff}.}

To explore this gap, we adopt visual-centric math as a representative task for complex visual reasoning and evaluate prominent pruning methods (i.e., FastV~\cite{FastV} and DivPrune~\cite{DivPrune}) with state-of-the-arts MLLMs~\cite{qwen3vl, internvl3_5} on visual math reasoning (VMR) benchmarks. 
Furthermore, we also evaluate these methods on simple VQA for comparison.\footnote{\label{note:vqa_vmr}\scriptsize VQA include SQA~\cite{ScienceQA}, GQA~\cite{GQA}, and \mm{\text{VQA}^{\text{T}}}~\cite{TextVQA}, and VMR include MathVerse~\cite{mathverse}, WeMath~\cite{wemath}, and DynaMath~\cite{dynamath}. \cref{fig:intro}(a) shows the average performance normalized to vanilla model as 1.0.}
As shown in~\cref{fig:intro}(a), we observe that while pruning methods effectively preserve performance on VQA, they exhibit precipitous performance drops on VMR. This consistent decline on VMR suggests that existing pruning methods fundamentally struggle to generalize to complex visual reasoning.

To uncover why pruning methods that succeed in VQA fail to generalize to VMR, we compare the visual attention map from the prefill stage (\cref{fig:intro}(b)) with those from key decoding steps (\cref{fig:intro}(c)–(f)), visualizing how the model’s focus transitions across the image throughout the decoding process.\footnote{\label{note:vqa_map}\scriptsize The corresponding attention visualizations for VQA benchmarks are provided in the~\cref{app:B_vqa_heatmap}.} 
Our visualization reveals a dynamic shift as reasoning progresses, demonstrating that the model’s visual focus does not remain fixed on the regions identified during the prefill stage; rather, it frequently transitions to entirely different visual areas to align with shifting reasoning requirements.
We term this dynamic shift of visual focus during decoding stage the \textit{\textbf{Relevant Visual Information Shift (RVIS)}}, indicating that the visual evidence required to resolve a task changes over time—a behavior fundamentally distinct from the static visual focus observed in VQA tasks.\textsuperscript{\ref{note:vqa_map}}

In this regard, we hypothesize that the failure of existing pruning methods in VMR stems from the occurrence of RVIS during the decoding stage, which their prefill-stage pruning strategies inherently struggle to address.
To validate this hypothesis, we conduct a detailed examination of the attention patterns of MLLMs in~\cref{sec:analysis} to uncover the root cause of the observed performance degradation.
Our diagnostic analysis yields two primary insights:
\textbf{1)} Visual reasoning exhibits drastic and frequent fluctuations in Relevant Visual Information throughout the decoding process, whereas such shifts are not observed in visual understanding.
\textbf{2)} This shift, which we term RVIS, serves as the primary driver of pruning failure, with performance declining sharply as its frequency increases during decoding.
Consequently, these observations underscore the necessity for an adaptive pruning approach during the decoding stage to effectively address RVIS.

To this end, we introduce \textbf{D}ecoding-stage \textbf{S}hift-aware \textbf{T}oken \textbf{P}runing (\proposed{}), a training-free and simple add-on to existing pruning methods.
{It adaptively restores previously discarded tokens by detecting RVIS in the model's visual focus during decoding, thereby aligning the visual tokens with the reasoning step.}
For this, \proposed{} preserves the tokens discarded by existing pruning methods for potential retrieval during decoding.
% Unlike existing methods that permanently discard tokens, \proposed{} keeps pruned tokens in a reserved set for potential retrieval in decoding. 
This mechanism is driven by two integrated modules: 
the \textbf{Relevant Visual Information Shift Detect (RISD)}, which monitors visual attention during decoding to identify RVIS through a threshold-based mechanism;
and the {\textbf{Context-Preserving Visual Token Swap (CPTS)}, which swaps current visual tokens with the newly relevant ones required by the model at that specific decoding step, thereby preventing erroneous generation.}
This simple \textit{Detect-and-Swap} design enables stable and context-aware reasoning while largely preserving the efficiency gains of conventional pruning methods.

Through extensive experiments, we demonstrate that \proposed{} significantly mitigates the performance degradation of conventional pruning in visual reasoning tasks, 
% by adaptively adjusting visual information,
while sustaining or marginally surpassing competitive performance on VQA tasks. 
Notably, these gains are achieved with minimal computational overhead, largely preserving the efficiency benefits of token pruning. 

We summarize our contributions as follows: 
\textbf{1)} We identify, for the first time, \textbf{RVIS} as a primary failure driver for existing pruning methods, particularly in reasoning-heavy tasks.
\textbf{2)} We propose \proposed{}, a training-free and simple add-on framework that leverages the RISD and CPTS modules to effectively address RVIS.
\textbf{3)} Extensive experiments demonstrate that \proposed{} significantly enhances performance in reasoning tasks with minimal computational overhead, proving its effectiveness and robustness in complex reasoning scenarios.

\section{Preliminary}
\label{sec:preliminary}
\textbf{Multimodal Large Language Models (MLLMs).}
Following the LLaVA~\cite{llava} design, MLLMs typically project an image and a text query into a shared token space. 
Let $X_v \in \mathbb{R}^{N_v \times d}$ and $X_t \in \mathbb{R}^{N_t \times d}$ denote the visual tokens processed through a vision encoder and text tokens, where $N_v$ and $N_t$ represent the number of visual and text tokens, and $d$ is the embedding dimension.
The LLM processes the concatenated token sequence $[X_v, X_t] \in \mathbb{R}^{(N_v + N_t) \times d}$ to generate a response.

% \subsection{Visual Token Pruning in MLLMs}
\smallskip \noindent  
\textbf{Visual Token Pruning in MLLMs.}
The goal of visual token pruning is to select the $k$ most relevant tokens from $X_v$ to reduce computational overhead, where $k$ denotes the token budget (i.e., the number of retained tokens). 
{To quantify the relevance of each visual token, the general attention weight vector $a_{t_l \to v} \in \mathbb{R}^{N_v}$ at any given decoding step $l$ is defined as follows:}
\begin{equation}
\label{eq:attn_vec}
a_{t_l \to v} = \text{Softmax}\left(q_{t_l} K_v^T /\sqrt{d_m}\right)
\end{equation}
where $q_{t_l} \in \mathbb{R}^{d_m}$ is the \textit{query} embedding of generated text token at step $l$, and $K_v \in \mathbb{R}^{N_v \times d_m}$ represents the \textit{key} embeddings of the entire visual token set $X_v$.

Conventional LLM-based pruning methods~\cite{FastV, PDrop} typically perform the pruning process as follows:
\textbf{Prefill Stage (\mm{l=0}):} In this initial stage, relevant scores are extracted at step $l=0$, where the query $q_{t_0}$ corresponds to the last token of the text sequence $X_t$ (i.e., the last instruction token).
{Based on the initial attention vector $a_{t_0 \to v}$, the pruned visual token set $X^{\text{Pruned}}_{v} \in \mb{R}^{k \times d_m}$ under token budget \mm{k} is then formed by selecting top-\mm{k} tokens associated with the highest attention weights:}
\begin{equation} \label{eq:Pruned_Set}
X_v^{\text{Pruned}} = X_v[\mathcal{I}],
\,\,\,\,\,
\mathcal{I} = \operatorname{TopK}\!\left(a_{t_0 \to v},\, k\right) 
\end{equation}
where $\operatorname{TopK}(a,k)$ selects the $k$ largest elements in vector $a$.
Subsequently, the reduced sequence $[X^{\text{Pruned}}_{v}, X_t]$ is fed into the LLM to initiate the decoding process.
\textbf{Decoding Stage (\mm{l > 0}):} In the decoding phase, the model sequentially generates tokens.
Let $G_l = \{g_1, g_2, \dots, g_l\}$ denote the tokens generated up to step $l$. At step $l+1$, the next token $g_{l+1}$ is predicted based on the pruned visual tokens \mm{X^{\text{Pruned}}_v} and the accumulated sequence \mm{X_t} and \mm{G_l}:
\begin{equation}
g_{l+1} = \text{LLM}(X^{\text{Pruned}}_{v}, X_t, G_l)
\end{equation}
Notably, existing pruning methods~\cite{FastV, VisionZip, DivPrune} permanently discard any visual tokens excluded from $X^{\text{Pruned}}_{v}$ during the prefill stage. 

\section{Why Does Visual Token Pruning Fail at Visual Reasoning?} \label{sec:analysis}
This section provides a diagnostic analysis to elucidate the performance disparity between VQA and VMR under existing pruning methods through the lens of the \textbf{\textit{Relevant Visual Information Shift (RVIS)}}. Our investigation is organized into two stages. First,\textbf{~\cref{sec:identify_key_factor}} identifies the existence of \textit{RVIS} during MLLM inference and characterizes its reasoning-intrinsic nature.
Second,\textbf{~\cref{sec:relation_rvis}} establishes \textit{RVIS} as the primary driver for pruning failure. We demonstrate that performance degradation stems directly from the conflict between these \textit{dynamic} shifts and \textit{static} pruning mechanisms of existing pruning methods, underscoring the critical necessity for an adaptive pruning strategy during the decoding stage.

\subsection{\textit{Relevant Visual Information Shift} as a Hallmark of Visual Reasoning}
\label{sec:identify_key_factor}
\begin{figure}[t]
    \centering
    \includegraphics[width=0.90\linewidth]{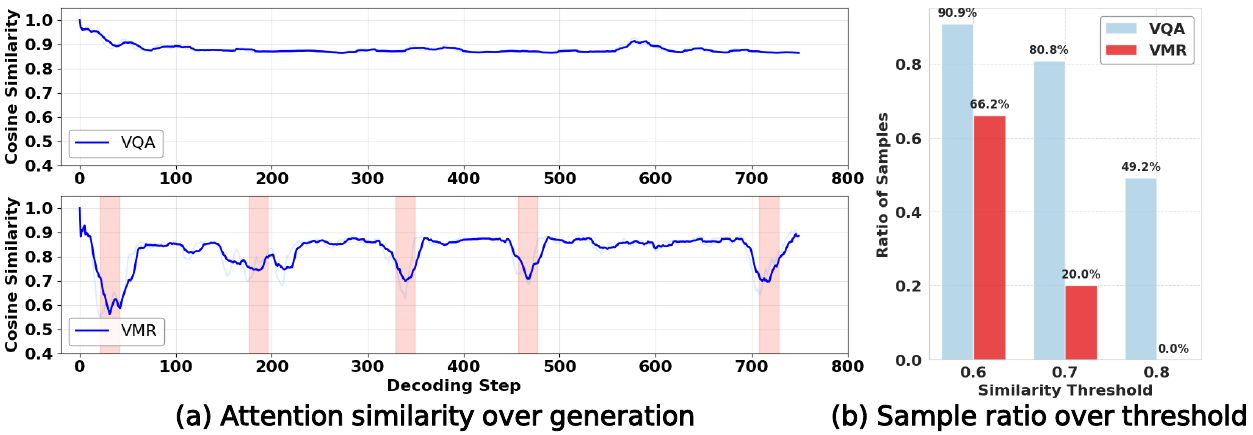}
    \caption{\small (a) Cosine similarity of visual attention distributions between the prefill stage ($l=0$) and each decoding step. (b) Proportion of samples maintaining attention similarity above thresholds throughout the entire decoding process.
    }
    \label{fig:VQA_VMR_Similarity}
\end{figure}

In this section, we investigate the fundamental attention dynamics that distinguish \textbf{visual reasoning} from \textbf{visual understanding} in MLLMs, adopting VMR and VQA as representative tasks.\footnote{\scriptsize We employ SQA~\cite{ScienceQA} and MathVerse~\cite{mathverse} as representative benchmarks for VQA and VMR tasks.}
Inspired by the qualitative observations in~\cref{fig:intro}(c)-(f), we hypothesize that attention stability, the degree to which a model maintains its initial visual focus thereafter, serves as a defining hallmark that differentiates these two abilities.
To validate this, we perform an attention-based analysis to define \textbf{\textit{Relevant Visual Information Shift (RVIS)}} and characterize its behavior.
{By tracking attention transitions throughout generation, we establish RVIS as the natural phenomenon of visual reasoning, where the model re-focuses on different visual regions as it progresses through sequential logical steps.}

\smallskip \noindent
\textbf{\textit{Relevant Visual Information Shift} in MLLMs.}
To provide a rigorous examination of the visual focus transitions observed earlier, we investigate the internal visual attention dynamics during the decoding process. 
{Specifically, we quantify the stability of the initial visual focus by calculating the cosine similarity between the visual attention vector at the prefill stage (\mm{a_{t_0 \to v}}) and those at each decoding step $l$ (\mm{a_{t_l \to v}}), denoted as $Sim(a_{t_0 \to v}, a_{t_l \to v})\in\mb{R}^{1}$.}

{As illustrated in~\cref{fig:VQA_VMR_Similarity}(a), we observe a stark contrast in stability between these two tasks. For VQA, the model maintains consistently high similarity, suggesting that it continues to attend to the regions prioritized during the prefill stage. In contrast, VMR exhibits sharp declines in similarity, revealing a transition toward different visual areas to satisfy new informational requirements. We formally define this phenomenon—where the model's focus deviates from the initial focus during the decoding—as \textbf{\textit{Relevant Visual Information Shift (RVIS)}}.} 
% As illustrated in Figure~\cref{fig:VQA_VMR_Similarity}(a), while VQA maintains consistently high similarity throughout generation, VMR exhibits sharp declines.
Notably,~\cref{fig:intro}(c)-(f) corresponds exactly to the point where RVIS is observed, confirming that visual reasoning entails a dynamic process where the necessary visual cues transition across successive reasoning steps.
Furthermore, we extended our analysis to a dataset-scale to verify the generalizability of RVIS. In~\cref{fig:VQA_VMR_Similarity}(b), we measure how many samples maintain their initial attention focus throughout the entire decoding process across different similarity thresholds.
In VMR, samples are significantly less likely to remain anchored to their prefill attention compared to VQA.
This gap validates that in reasoning-intensive tasks, visual relevance is inherently dynamic, necessitating substantial shifts as the model progresses through logical steps.

\begin{wrapfigure}{r}{0.44\linewidth}
    \centering
    \vspace{-5pt}
    \includegraphics[width=0.99\linewidth]{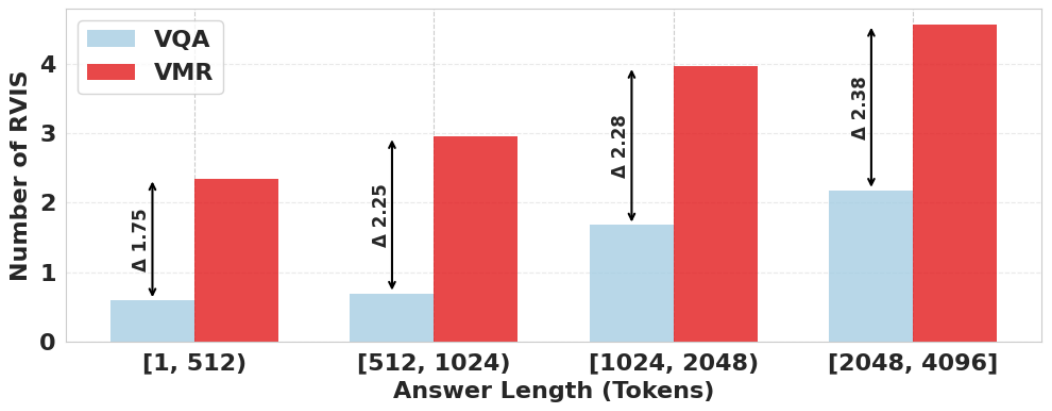}
    \vspace{-4ex}
    \caption{\small Average number of RVIS occurrences across various answer lengths.
    % in ScienceQA (VQA) and MathVerse (VMR).
    }
    \vspace{-5pt}
    \label{fig:rvis_occurrence}
\end{wrapfigure}

\smallskip \noindent \textbf{Reasoning-Intrinsic Nature of RVIS.} 
To validate that RVIS is intrinsically driven by the reasoning process, not merely by the extended generation lengths typical of VMR, we analyze its occurrence frequency across controlled sequence length intervals.\footnote{\scriptsize To identify RVIS, we detect similarity drops below a threshold of \textit{0.7}. To ensure each RVIS event is counted as a single occurrence rather than multiple consecutive steps below the threshold, we utilize the \texttt{find\_peaks} algorithm on the similarity signal. Further details are provided in~\cref{app:C_find_peak}.}
As shown in~\cref{fig:rvis_occurrence}, the frequency of RVIS remains consistently high in VMR regardless of response length. Notably, even the shortest VMR samples (1-512 tokens) exhibit a higher RVIS frequency than the longest VQA samples (2048-4096 tokens).
{This disparity confirms RVIS as an inherent hallmark of the reasoning process, rather than a mere side effect of generation length.
In this process, visual focus dynamically shifts to align with each successive logical step.
We provide additional experiments on the reasoning-intrinsic nature of RVIS in \cref{app:D_RVIS_Nature_additional}.
}

\smallskip \noindent \ul{\textbf{Discussion.}} 
Crucially, RVIS is not a manifestation of model failure but a natural behavior of MLLMs navigating multi-step reasoning. Acknowledging RVIS as a systemic property establishes the necessity for adaptive visual token pruning during decoding when applying pruning methods in the following section.

\subsection{How do Relevant Visual Information Shift affect token pruning performance?}
\label{sec:relation_rvis}
Building on our understanding of RVIS, we shift our focus toward its direct impact on existing pruning methods. 
We argue that the performance discrepancy observed in~\cref{fig:intro}(a) stems from a fundamental conflict: the \textit{dynamic} informational needs of RVIS versus the \textit{static} pruning mechanisms inherent to existing pruning methods.
% we re-evaluate existing pruning methods through the lens of RVIS to explain the performance discrepancy observed in Figure~\cref{fig:intro}(a). 
% We demonstrate that the primary bottleneck stems from a fundamental conflict: the \textit{dynamic} informational needs of RVIS versus the \textit{static} pruning mechanism inherent to conventional approaches.

\smallskip \noindent
\textbf{Task-Specific Distribution of RVIS Frequency.}
We first examine the prevalence of RVIS by quantifying occurrences per sample to contrast the divergent RVIS frequency patterns between VQA and VMR.\footnote{\scriptsize We initially sampled 300 instances per task. After filtering those truncated by the 4,096 \texttt{max\_generation\_len}, we balanced the datasets to match the smaller group, resulting in $N=276$ samples for each task.}
We first examine the prevalence of RVIS across VQA and VMR. 

\begin{wrapfigure}{r}{0.44\linewidth}
    \centering
    \vspace{-25pt}
    \includegraphics[width=0.99\linewidth]{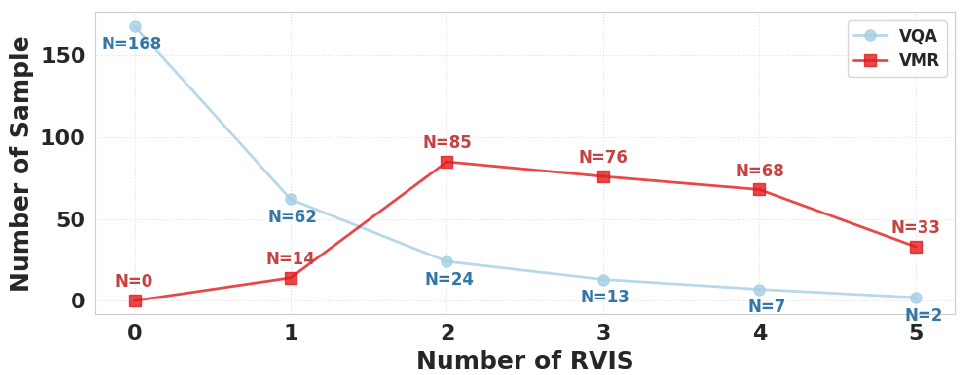}
    \vspace{-3ex}
    \caption{\small Distribution of RVIS occurrences for VQA and VMR. $N$ indicates the sample count for each bin.
    }
    \vspace{-23pt}
    \label{fig:rvis_distribution}
\end{wrapfigure}
\cref{fig:rvis_distribution} reveals a stark contrast: while VQA samples are largely static, with the vast majority exhibiting zero or minimal RVIS, VMR samples exhibit two or more shifts during the decoding.
This re-confirms that RVIS is fundamentally driven by reasoning intensity. As task complexity increases, the MLLM must transition its visual focus to gather the necessary evidence for each sequential logical step.

\begin{wrapfigure}{r}{0.44\linewidth}
    \centering
    \vspace{-12pt}
    \includegraphics[width=0.99\linewidth]{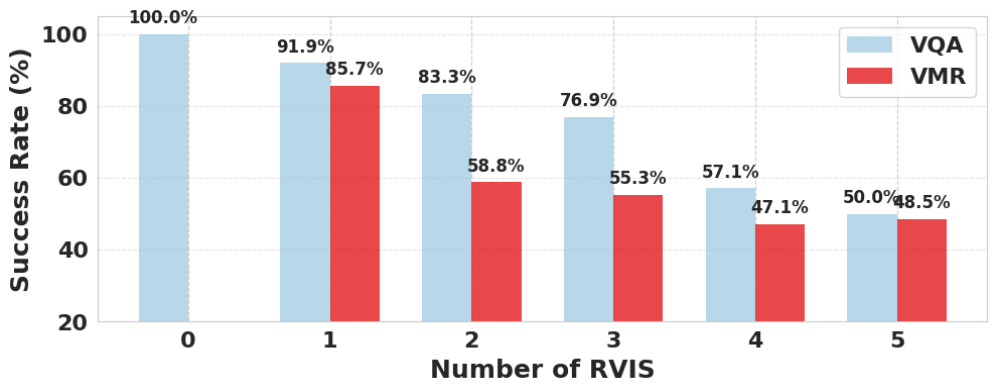}
    \vspace{-4ex}
    \caption{\small Success rate of FastV~\cite{FastV} across different RVIS frequencies.
    }
    \vspace{-22pt}
    \label{fig:rvis_causal}
\end{wrapfigure}
\smallskip \noindent
\textbf{Impact of RVIS on Pruning Success.}
To establish the direct impact of RVIS when combined with existing pruning methods, we analyze the pruning success rate across varying RVIS frequencies. 
For this, we define the success rate as the ratio of samples correctly solved after pruning relative to those solved in vanilla model which utilizes the entire visual token set without pruning.
As illustrated in~\cref{fig:rvis_causal}, the success rate of pruning methods drops precipitously with increasing RVIS frequency, a robust trend 
that holds for both tasks.
% across diverse tasks.
% observed in both VQA and VMR tasks. 
% regardless of tasks.
This provides direct evidence that the failure of existing pruning methods is rooted in their inability to adapt to the model's shifting visual focus during decoding.
Specifically, \textbf{\textit{when}} RVIS occurs—requiring visual information that deviates from that identified during the initial prefill stage—static pruning methods that discard tokens at the prefill become inherently prone to failure.
This illustrates exactly \textbf{\textit{why}} existing pruning methods struggle with reasoning-heavy tasks like VMR, as observed in~\cref{fig:intro}(a).

\smallskip
In summary, our analysis identifies the Relevant Visual Information Shift (RVIS) as the primary failure driver that limits the efficacy of existing \textit{static} pruning methods, highlighting the need for an \textit{adaptive} framework that can update the visual tokens as the MLLM's focus transitions during decoding.

\section{\proposed{}: \textbf{D}ecoding-stage \textbf{S}hift-aware \textbf{T}oken \textbf{P}runing}
Building on our analysis in~\cref{sec:analysis}, we propose \textbf{D}ecoding-stage \textbf{S}hift-aware \textbf{T}oken \textbf{P}runing (\proposed{}), a simple yet effective add-on framework 
designed to rectify the limiatation of existing static pruning methods by adaptively updating visual tokens during decoding.
To this end, \proposed{} integrates \textbf{Relevant Visual Information Shift Detect (RISD)} to detect the onset of RVIS throughout the decoding process (\cref{sec:RISD}). This signal triggers \textbf{Context-Preserving Visual Token Swap (CPTS)} to retrieve newly relevant visual tokens, guiding MLLM to re-focus the necessary visual cues at that specific decoding steps (\cref{sec:CPTS}). The overall framework is shown in~\cref{fig:overall_arch}.\footnote{\scriptsize The detailed algorithm for \proposed{} is in~\cref{app:E_method_algorithm}.}

\subsection{Relevant Visual Information Shift Detect (RISD)} \label{sec:RISD}

\begin{figure}[t]
    \centering
    \includegraphics[width=0.9\linewidth]{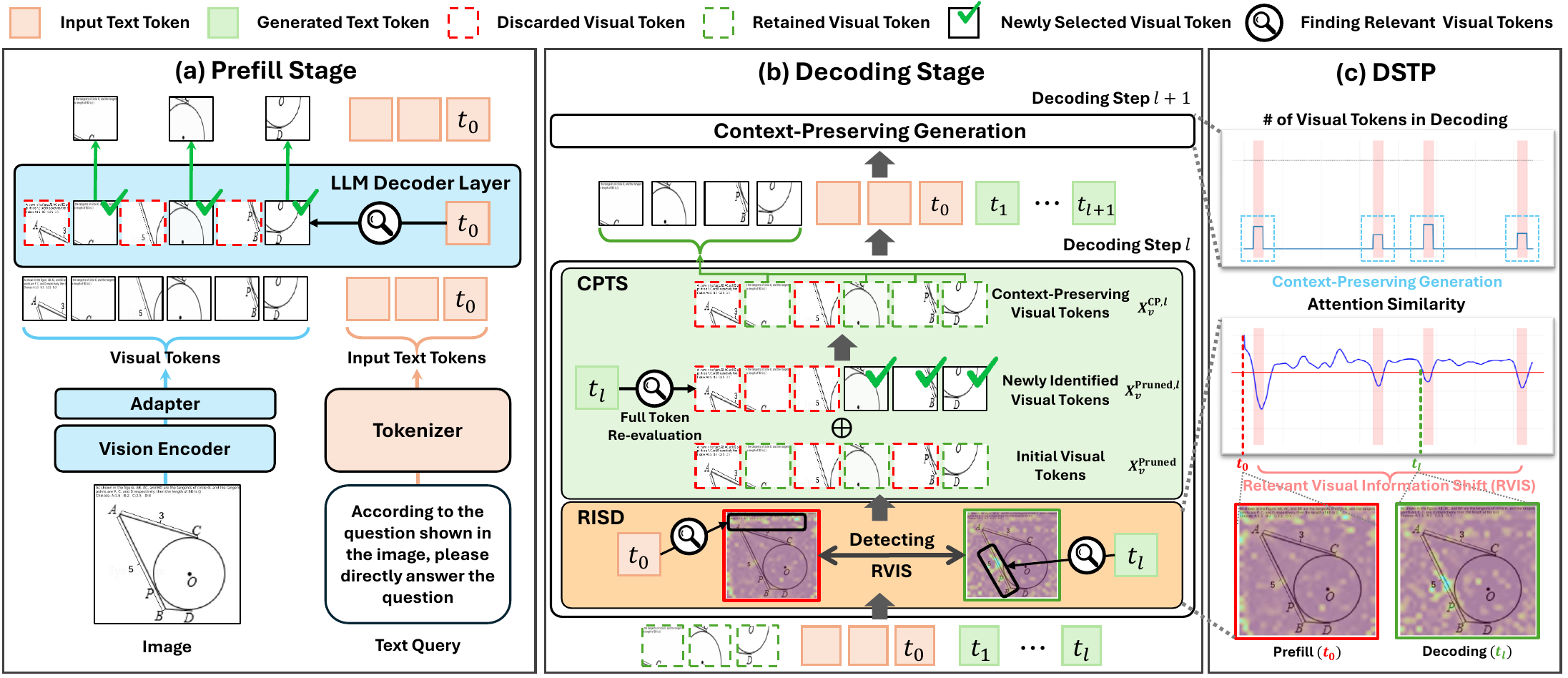}
    \caption{\small Overall framework of \proposed{}. (a) Prefill-stage protocol of base pruning methods. (b) RISD and CPTS modules at Decoding-stage. RISD monitors attention similarity at each decoding step, invoking CPTS for context-preserving token swapping when RVIS is detected. (c) Overall flow of \proposed{} throughout the entire decoding process.}
    \label{fig:overall_arch}
\end{figure}

\noindent
\textbf{Prefill-stage Setup.}
\proposed{} follows the original prefill-stage protocol of base pruning methods~\cite{FastV, SparseVLM}, which is illustrated in~\cref{fig:overall_arch}(a), as its core functionality is designed to operate during decoding. To enable RVIS detection, the RISD module first extracts the initial attention vector $a_{t_0 \to v} \in \mathbb{R}^{N_v}$ during the prefill stage.
This vector, which is used to determine the important visual tokens in~\cref{eq:attn_vec}, serves as the anchor attention to represent the model's initial visual focus.
{Also, we define the \textit{Reserved Visual Tokens Set} $X^{\text{Reserved}}_{v} = X_v - X^{\text{Pruned}}_v \in \mb{R}^{(N_v - k) \times d_m}$, which consists of the tokens discarded but kept in standby for potential recovery.}

\smallskip \noindent
\textbf{Decoding-stage Monitoring.} 
{As decoding progresses, the MLLM may require visual cues that deviate from the initial focus captured by the anchor attention vector \mm{a_{\textcolor{black}{t_0} \to v}} at prefill-stage (\textcolor{red}{\mm{l=0}}).
Based on our analysis in~\cref{sec:analysis}, we argue that RVIS characterizes these informational needs; 
therefore, as depicted in the RISD module of \cref{fig:overall_arch}(b), we track the current visual focus (\textcolor{Green}{\mm{t_l}}) and compare it with the initial visual focus (\textcolor{red}{\mm{t_0}}) at each decoding step to detect its onset.
Specifically, we monitor the attention vector $a^{\text{Pruned}}_{\textcolor{black}{t_l} \to v}$ at each decoding \textcolor{Green}{step $l$}:}
\begin{equation}
a_{\textcolor{black}{t_l} \to v}^{\text{Pruned}} = \text{Softmax} ( q_l (K^{\text{Pruned}}_v)^\top / \sqrt{d_m} ) \in \mb{R}^{k}
\label{eq:inst_attn}
\end{equation}
{where \mm{k} is the token budget, $q_l \in \mathbb{R}^{d_m}$ is the \textit{query} embedding of generated text token at step $l$, and $K^{\text{Pruned}}_v\in \mb{R}^{k \times d_m}$ denotes the \textit{key} embedding of the visual tokens set \mm{X^{\text{Pruned}}_v} retained by base pruning methods during the prefill stage.
To quantify the alignment between the model's current visual focus and the initial anchor, we calculate the cosine similarity $s_l \in \mb{R}$ at each decoding step \mm{l}:}
\begin{equation} \label{eq:5_sim}
s_l = \text{Sim}(\textcolor{red}{a^{\text{Pruned}}_{t_0 \to v}}, \textcolor{Green}{a^{\text{Pruned}}_{t_l \to v}}) = \frac{a^{\text{Pruned}}_{t_0 \to v} \cdot a^{\text{Pruned}}_{t_l \to v}}{|a^{\text{Pruned}}_{t_0 \to v}| |a^{\text{Pruned}}_{t_l \to v}|}
\end{equation}
As illustrated in \cref{fig:overall_arch}(c),
% (\textcolor{red}{\mm{t_0}} versus \textcolor{green}{\mm{t_l}})
a decline in $s_l$ indicates an attention shift toward visual regions that are different from the initial focus, signaling the onset of RVIS. 
We utilize a constant threshold $\tau$ to detect RVIS, as it provides a computationally lightweight yet robust trigger, performing comparably to more complex divergence metrics as demonstrated in~\cref{sec:indepth_analysis}.
Once $s_l < \tau$ is detected, the CPTS module is immediately invoked to re-evaluate and retrieve the relevant visual tokens based on the current text query.

\subsection{Context-Preserving Visual Token Swap (CPTS)} \label{sec:CPTS}
% \vspace{-0.4ex}
\noindent
\textbf{Dynamic Importance Re-evaluation.}
To pinpoint the precise visual tokens required for the current decoding step, it is essential to consider the complete set $X_v$, including those previously discarded by the baseline pruning method during the prefill stage.
Consequently, we perform a comprehensive re-evaluation of $X_v$ with respect to the current text query $q_l$.
Specifically, we leverage the attention vector $a^{\text{Pruned}}_{t_l \to v}$, which is internally computed as part of LLM decoding process.
To assess the discarded visual tokens, we additionally compute $a^{\text{Reserved}}_{t_l \to v}$ for $X^{\text{Reserved}}_v$ using~\cref{eq:inst_attn}.
By concatenating these weights, we construct the full importance score across the entire visual token set:
\begin{equation} \label{eq:full_attn_CPTS}
a_{t_l \to v} = \left[ a^{\text{Pruned}}_{t_l \to v}, a^{\text{Reserved}}_{t_l \to v} \right] \in \mathbb{R}^{N_v}
\end{equation}
This approach enables a full importance scoring across $X_v$ with minimal computational overhead, allowing the selection of a newly relevant top-$k$ visual token set, $X^{\text{Pruned}, l}_v \in \mathbb{R}^{k \times d_m}$, based on the updated scores $a_{t_l \to v}$.
For instance, let $v_1, \dots, v_6$ denote the visual tokens in the CPTS module of~\cref{fig:overall_arch}(b) from left to right. 
In this case, ${X}^{\text{Pruned}, l}_v$ identifies $\{v_4, v_5, v_6\}$ as newly relevant, which differs from the prefill-stage set ${X}^{\text{Pruned}}_v$ consisting of $\{v_2, v_4, v_6\}$. 
This ensures that the MLLM retrieves the most appropriate visual tokens in alignment with the current decoding steps.

\smallskip \noindent
\textbf{Context-Preserving Visual Token Swap.} 
{Visual tokens provide the essential image context that allows MLLMs to perform a wide range of multimodal tasks.
However, naively replacing the initial visual token set $X^{\text{Pruned}}_v$ with a newly identified set $X^{\text{Pruned}, l}_v$ can hinder the model’s ability to maintain a stable understanding of the image, as there is little semantic overlap between the two sets.
Such a sudden context shift disrupts the model's internal consistency, leading to erroneous outputs.
To address this, we introduce the \textit{Context-Preserving Visual Token Swap}. 
Rather than an complete replacement, we merge the initial prefill-stage visual tokens \mm{X^{\text{Pruned}}_v} with the newly relevant tokens \mm{X^{\text{Pruned}, l}_v} through a union operation to form an unified (context-preserving) visual token set $X^{CP, l}_v$:}
\begin{equation}
X^{CP, l}_v = X^{\text{Pruned}}_v \cup X^{\text{Pruned}, l}_v \in \mathbb{R}^{k^* \times d_m}
\label{eq:union}
\end{equation}
\textcolor{black}{where $k^*$ denotes the size of the union set ($k \le k^* \le 2k$).
In this scenario, ${X}^{CP, l}_v$ is $\{v_2, v_4, v_5, v_6\}$, where the newly added $v_5$ guides the current reasoning step while the remaining tokens sustain the original visual context.
% Here, the newly incorporated token $v_5$ guides the model's reasoning by providing the updated visual cues required for the current step, while the original tokens maintain the overall image context.
To ensure the MLLM effectively incorporates these new tokens, we maintain this temporarily expanded set for a context-preserving duration of $L$ decoding steps as illustrated in~\cref{fig:overall_arch}(c).
After this duration, the initial prefill-stage set $X^{\text{Pruned}}_v$ is reused as the visual context to minimize additional re-evaluation overhead. Furthermore, as the model's focus typically aligns with the initial anchor in the absence of RVIS, as shown in~\cref{sec:analysis}, $X^{\text{Pruned}}_v$ remains sufficient for the subsequent decoding steps.\footnote{\scriptsize Detailed analysis of this reversion strategy is provided in~\cref{app:F_Token_Reversion}.}
This transient increase in the number of visual tokens during the context-preserving duration does not incur long-term computational inefficiency, as confirmed by our analysis in~\cref{sec:indepth_analysis}.}

\section{Experiment}
\subsection{Experimental Settings}
\renewcommand{\arraystretch}{0.85}
\begin{table*}[t]
\centering
\caption{\small \textbf{Visual Reasoning} Performance of \textbf{Qwen3-VL-4B} and \textbf{InternVL3.5-8B} under different token retention ratios with various pruning methods and \proposed{}. Values in parentheses are the proportion relative to the upper bound.}
\label{tab:qwen_visual_reasoning}
\setlength{\tabcolsep}{2.5pt}
\resizebox{0.87\linewidth}{!}{
\begin{tabular}{c|ccccc|c}

\specialrule{0.15em}{0em}{0em}
\multicolumn{7}{c}{\textbf{\textit{Qwen3-VL-4B}}~\cite{qwen3vl}} \\
\specialrule{0.15em}{0em}{0em}
\textbf{Method} & \textbf{MathVerse} & \textbf{WeMath} & \textbf{DynaMath} & \textbf{LogicVista} & \textbf{MMMU-Pro} &\textbf{Acc. (\%)} \\ \specialrule{0.1em}{0em}{0em}
% \rowcolor{gray!20} \multicolumn{7}{c}{\textit{Upper Bound, Retain Full Visual Tokens (100\%)}} \\
Vanilla (Full Tokens) & 61.29 &  48.29 & 66.48 & 49.22 & 37.63 & 100\% \\ \specialrule{0.1em}{0em}{0em}

\rowcolor{gray!20} \multicolumn{7}{c}{\textit{Retain 33.3\% Tokens ($\downarrow$ 66.7\%)}} \\ 

FastV \scriptsize(ECCV'24) & 32.23 \scriptsize{(52.6\%)}  & 25.90 \scriptsize{(53.6\%)} & 38.76 \scriptsize{(58.3\%)}  & 30.64 \scriptsize{(62.3\%)} & 19.41 \scriptsize{(51.6\%)} & 55.68\% \\

w/ \proposed{} & 52.54 \scriptsize{(85.7\%)} & 42.19 \scriptsize{(87.4\%)} & 51.00 \scriptsize{(76.7\%)} & 41.16 \scriptsize{(83.6\%)} & 30.52 \scriptsize{(81.1\%)} & 82.90\% \\ \midrule

DivPrune \scriptsize(CVPR'25) & 33.90 \scriptsize{(55.3\%)} & 29.32 \scriptsize{(60.7\%)} & 41.94 \scriptsize{(63.1\%)} & 30.64 \scriptsize{(62.3\%)} & 13.08 \scriptsize{(34.8\%)} & 55.24\% \\

w/ \proposed{} & 50.25 \scriptsize{(82.0\%)} & 42.95 \scriptsize{(88.9\%)} & 57.70 \scriptsize{(86.8\%)} & 42.70 \scriptsize{(86.8\%)} & 28.03 \scriptsize{(74.5\%)} & 83.80\% \\ \midrule

VisionZip \scriptsize(CVPR'25) & 36.80 \scriptsize{(60.0\%)} & 35.57 \scriptsize{(73.7\%)} & 43.16 \scriptsize{(64.9\%)} & 29.53 \scriptsize{(60.0\%)} & 16.36 \scriptsize{(43.5\%)} & 60.42\% \\

w/ \proposed{} & 50.76 \scriptsize{(82.8\%)} & 43.62 \scriptsize{(90.3\%)} & 52.95 \scriptsize{(79.6\%)} & 39.22 \scriptsize{(79.7\%)} & 27.39 \scriptsize{(72.8\%)} & 81.04\% \\ \specialrule{0.1em}{0em}{0em}

%% Retain 22.2%
\rowcolor{gray!20} \multicolumn{7}{c}{\textit{Retain 22.2\% Tokens ($\downarrow$ 77.8\%)}} \\ 
FastV \scriptsize(ECCV'24) & 25.88 \scriptsize{(42.2\%)}  & 21.43 \scriptsize{(44.4\%)} & 35.85 \scriptsize{(53.9\%)}  & 26.96 \scriptsize{(54.8\%)} & 11.50 \scriptsize{(30.6\%)} & 45.18\% \\

w/ \proposed{} & 42.76 \scriptsize{(69.8\%)} & 31.71 \scriptsize{(65.7\%)} & 47.31 \scriptsize{(71.2\%)} & 39.05 \scriptsize{(79.3\%)} & 26.87 \scriptsize{(71.4\%)} & 71.48\% \\ \midrule

DivPrune \scriptsize(CVPR'25) & 26.14 \scriptsize{(42.6\%)} & 24.18 \scriptsize{(50.1\%)} & 38.36 \scriptsize{(57.7\%)} & 27.96 \scriptsize{(56.8\%)} & 11.21 \scriptsize{(29.8\%)} & 47.40\% \\

w/ \proposed{} & 46.19 \scriptsize{(75.4\%)} & 40.00 \scriptsize{(82.8\%)} & 52.29 \scriptsize{(78.7\%)} & 35.55 \scriptsize{(72.2\%)} & 21.66 \scriptsize{(57.6\%)} & 73.34\% \\ \midrule

VisionZip \scriptsize(CVPR'25) & 25.98 \scriptsize{(42.4\%)} & 23.81 \scriptsize{(49.3\%)} & 35.90 \scriptsize{(54.0\%)} & 25.50 \scriptsize{(51.8\%)} & 13.00 \scriptsize{(34.5\%)} & 46.40\% \\

w/ \proposed{} & 40.46 \scriptsize{(66.0\%)} & 34.19 \scriptsize{(70.8\%)} & 47.48 \scriptsize{(71.4\%)} & 37.58 \scriptsize{(76.4\%)} & 22.25 \scriptsize{(59.1\%)} & 68.74\% \\ \specialrule{0.1em}{0em}{0em}

% \end{tabular}
% }
% \vspace{-3ex}
% \end{table*}

% \begin{table*}[h]
% \centering
% \caption{\textbf{Visual Reasoning} Performance of \textbf{InternVL3.5-8B} under different pruning ratios with various VTP methods and \proposed{}. Values in parentheses are the proportion relative to the upper bound.}
% \label{tab:intern_visual_reasoning}
% \vspace{-3ex}
% \resizebox{0.95\linewidth}{!}{
% \begin{tabular}{c|ccccc|c}

\specialrule{0.05em}{0em}{0em}
\multicolumn{7}{c}{\textbf{\textit{InternVL3.5-8B}}~\cite{internvl3_5}} \\
\specialrule{0.15em}{0em}{0em}

% \specialrule{0.1em}{0em}{0em}
\textbf{Method} & \textbf{MathVerse} & \textbf{WeMath} & \textbf{DynaMath} & \textbf{LogicVista} & \textbf{MMMU-Pro} & \textbf{Acc. (\%)} \\ \specialrule{0.1em}{0em}{0em}
% \rowcolor{gray!20} \multicolumn{7}{c}{\textit{Upper Bound, Full Visual Tokens (100\%)}} \\
Vanilla (Full Tokens) & 42.00 & 43.62 & 48.26 & 52.13 &  39.42 & 100.00\% \\ \specialrule{0.1em}{0em}{0em}

%% Retain 33.3%
\rowcolor{gray!20} \multicolumn{7}{c}{\textit{Retain 33.3\% Tokens ($\downarrow$ 66.7\%)}} \\ 
FastV \scriptsize(ECCV'24) & 21.57 \scriptsize{(51.4\%)} & 28.38 \scriptsize{(65.1\%)} & 26.71 \scriptsize{(55.3\%)} & 33.08 \scriptsize{(63.5\%)} & 18.96 \scriptsize{(48.1\%)} & 56.66\% \\

w/ \proposed{} & 40.22 \scriptsize{(95.8\%)} & 40.38 \scriptsize{(92.6\%)} & 45.38 \scriptsize{(94.0\%)} & 48.44 \scriptsize{(92.9\%)} & 34.27 \scriptsize{(86.9\%)} & 92.44\% \\ \midrule

DivPrune \scriptsize(CVPR'25) & 26.59 \scriptsize{(63.3\%)} & 32.62 \scriptsize{(74.8\%)} & 32.57 \scriptsize{(67.5\%)} & 32.78 \scriptsize{(62.9\%)} & 19.42 \scriptsize{(49.3\%)} & 63.55\% \\

w/ \proposed{} & 39.59 \scriptsize{(94.3\%)} & 41.71 \scriptsize{(95.6\%)} & 43.22 \scriptsize{(89.6\%)} & 47.87 \scriptsize{(91.8\%)} & 31.21 \scriptsize{(79.2\%)} & 90.09\% \\ \midrule

VisionZip \scriptsize(CVPR'25) & 24.84 \scriptsize{(59.1\%)} & 29.14 \scriptsize{(66.8\%)} & 30.54 \scriptsize{(63.3\%)} & 34.45 \scriptsize{(66.1\%)} & 16.71 \scriptsize{(42.4\%)} & 59.54\% \\
w/ \proposed{} & 37.56 \scriptsize{(89.4\%)} & 41.52 \scriptsize{(95.2\%)} & 43.11 \scriptsize{(89.3\%)} & 48.76 \scriptsize{(93.5\%)} & 28.26 \scriptsize{(71.7\%)} & 87.83\% \\ \specialrule{0.1em}{0em}{0em}

%% Retain 22.2%
\rowcolor{gray!20} \multicolumn{7}{c}{\textit{Retain 22.2\% Tokens ($\downarrow$ 77.8\%)}} \\ 

FastV \scriptsize(ECCV'24) & 20.06 \scriptsize{(47.8\%)} & 26.90 \scriptsize{(61.7\%)} & 24.29 \scriptsize{(50.3\%)} & 30.20 \scriptsize{(57.9\%)} & 11.60 \scriptsize{(29.4\%)} & 49.42\% \\

w/ \proposed{} & 38.94 \scriptsize{(92.7\%)} & 37.57 \scriptsize{(86.1\%)} & 42.95 \scriptsize{(89.0\%)} & 46.08 \scriptsize{(88.4\%)} & 31.04 \scriptsize{(78.7\%)} & 87.00\% \\ \midrule

DivPrune \scriptsize(CVPR'25) & 23.64 \scriptsize{(56.3\%)} & 30.81 \scriptsize{(70.6\%)} & 29.02 \scriptsize{(60.1\%)} & 34.00 \scriptsize{(65.2\%)} & 14.10 \scriptsize{(35.8\%)} & 57.61\% \\

w/ \proposed{} & 36.54 \scriptsize{(87.0\%)} & 38.86 \scriptsize{(89.1\%)} & 39.29 \scriptsize{(81.4\%)} & 44.74 \scriptsize{(85.8\%)} & 28.21 \scriptsize{(71.6\%)} & 82.98\% \\ \midrule

VisionZip \scriptsize(CVPR'25) & 20.55 \scriptsize{(48.9\%)} & 26.29 \scriptsize{(60.3\%)} & 26.34 \scriptsize{(54.6\%)} & 31.76 \scriptsize{(60.9\%)} & 12.02 \scriptsize{(30.5\%)} & 51.04\% \\

w/ \proposed{} & 34.13 \scriptsize{(81.3\%)} & 39.52 \scriptsize{(90.6\%)} & 40.51 \scriptsize{(83.9\%)} & 43.62 \scriptsize{(83.7\%)} & 25.85 \scriptsize{(65.6\%)} & 81.01\% \\ \specialrule{0.1em}{0em}{0em}

\end{tabular}
}
\vspace{-4ex}
\end{table*}
\textbf{Datasets.}
To validate the effectiveness of \proposed{} in reasoning-heavy scenarios, we categorize our benchmarks into two primary domains. First, we focus on \textbf{Visual Reasoning}, which is further subdivided into: (i) \textit{Visual Math Reasoning} (i.e., MathVerse~\cite{mathverse}, WeMath~\cite{wemath}, and DynaMath~\cite{dynamath}), requiring rigorous step-by-step mathematical derivation; and (ii) \textit{Puzzle and STEM-related Reasoning} (i.e., LogicVista~\cite{logicvista} and MMMU-Pro~\cite{mmmu-pro}), which assess the complex logical and expert-level analytical capabilities.
Furthermore, we evaluate standard \textbf{Visual Understanding} benchmarks, including GQA~\cite{GQA}, \mm{\text{VQA}^{\text{T}}}~\cite{TextVQA}, and SQA~\cite{ScienceQA}, to provide a comprehensive comparison with existing pruning literature~\cite{FastV, SparseVLM, VisionZip}.

\smallskip \noindent
\textbf{Baselines.}
To demonstrate the backbone-agnostic nature of \proposed{} as a plug-and-play framework, we evaluate its effectiveness by integrating it with representative visual pruning baselines: FastV~\cite{FastV}, DivPrune~\cite{DivPrune}, and VisionZip~\cite{VisionZip}.
We present the performance of each baseline with and without our method to highlight the improvements and validate its robustness across diverse pruning strategies.
Also, we provide additional generalizability experiments in~\cref{app:G_General_Experiment}.

\smallskip \noindent \textbf{Implementation Details.}
We implement our method on Qwen3-VL-4B~\cite{qwen3vl} and InternVL3.5-8B~\cite{internvl3_5}.\footnote{\scriptsize All experiments are conducted on NVIDIA L40S GPUs.} 
The detection threshold $\tau$ was set to \textit{0.75} and the context preserving duration $L$ was fixed at \textit{20}.
% Following the experimental setup in~\cite{FastV, SparseVLM}, we evaluate our method using token retention ratios of 33.3\% and 22.2\%. For all evaluations, the maximum generation length is set to 4,096 tokens. Unless otherwise stated, we conduct experiments using Qwen3-VL-4B as the primary backbone. To further demonstrate the generalizability of our framework, additional evaluations are provided in~\cref{app:G_General_Experiment}.
\textcolor{black}{Additional implementation details are provided in the~\cref{app:H_Implementation_Detail}.}

\subsection{Main Results}

\textbf{Visual Reasoning Benchmarks.}
In~\cref{tab:qwen_visual_reasoning}, we evaluate \proposed{} when integrated with diverse pruning methods across various visual reasoning benchmarks. Our analysis reveals the following key insights:
\textbf{1)} Existing static pruning methods suffer from a severe performance degradation across all reasoning benchmarks, regardless of the MLLM architecture.
This confirms that static pruning methods fundamentally struggle to address RVIS.
\textbf{2)} Integrating \proposed{} with these methods consistently overcomes such limitations by significantly improving reasoning performance.
This result demonstrates that our framework successfully addresses RVIS, ensuring MLLMs re-focus on the necessary visual tokens as informational needs transition during the reasoning process.
\textbf{3)}
While \proposed{} achieves consistent performance improvements across all evaluated datasets, the most significant gains are observed in benchmarks characterized by high visual complexity such as MathVerse and MMMU-Pro~\cite{ARES, mmmu-pro}.\footnote{\scriptsize A comprehensive analysis regarding the visually intensive group (MathVerse and MMMU-Pro) and other benchmarks is provided in the~\cref{app:I_intensive_analysis}} 
This performance gap demonstrates that \proposed{} is more effective in reasoning intensive tasks, where complex logical reasoning is essential.

\renewcommand{\arraystretch}{0.85}
\begin{wraptable}{r}{0.38\textwidth} 
    \centering
    \vspace{-36pt} % 상단 여백 조절
    \caption{\textbf{Visual Understanding} Performance of \textbf{Qwen3-VL-4B} with 33.3\% retention ratio.}
    \label{tab:visual_understanding}
    \setlength{\tabcolsep}{2.5pt}
    \resizebox{0.95\linewidth}{!}{
        \begin{tabular}{l|ccc|c}
        \specialrule{0.15em}{0em}{0em}
            \textbf{Method} & \textbf{SQA} & \textbf{\mm{\text{VQA}^{\text{T}}}} & \textbf{GQA} & \textbf{Acc.} \\ 
            \specialrule{0.1em}{0em}{0em}
            Vanilla & 93.42 & 81.57 & 61.82 & 100\% \\ \specialrule{0.1em}{0em}{0em}
            FastV  & 87.98 & 74.82 & 60.04 & 94.3\% \\
            {w/ \proposed{}} & {91.84} & {77.95} & {60.93} & {97.5\%} \\ \midrule
            DivPrune & 86.12 & 71.36 & 58.07 & 91.2\% \\
            {w/ \proposed{}} & {91.36} & {74.11} & {60.42} & {95.5\%} \\ \midrule
            VisionZip & 90.41 & 77.10 & 60.68 & 96.5\% \\
            {w/ \proposed{}} & {92.08} & {77.20} & {61.27} & {97.4\%} \\ 
            \bottomrule
        \end{tabular}
    }
    \vspace{-25pt} % 하단 여백 조절
\end{wraptable}

\smallskip \noindent
\textbf{Visual Understanding Benchmarks.}
We evaluate \proposed{} on visual understanding benchmarks using Qwen3-VL-4B, as summarized in~ \cref{tab:visual_understanding}.\footnote{\scriptsize For extended experimental results, please refer to \cref{app:J_VQA}.}
Our key observations are as follows:
% We evaluate \proposed{} on visual understanding benchmarks to ensure a fair and comprehensive comparison with existing pruning methods in~\cref{tab:visual_understanding} on Qwen3-VL-4B.\footnote{\scriptsize For more comprehensive experimental results, please refer to \cref{app:J_VQA}.}
% Our key observations are as follows:
\textbf{1)} 
Static methods exhibit relatively stable performance in visual understanding compared to reasoning tasks. This is because minimal RVIS in visual understanding tasks, as shown in~\cref{sec:analysis}, allows prefill-stage pruning to remain largely effective. Nevertheless, integrating \proposed{} consistently yields further performance gains. 
% Static pruning methods exhibit relatively stable performance on visual understanding tasks compared to reasoning tasks. This is because visual understanding typically involves minimal RVIS, as shown in~\cref{sec:analysis}, allowing the pruning made during the prefill stage to remain largely effective throughout the process.
\textbf{2)} 
The performance boost is particularly significant on SQA. Unlike GQA or \mm{\text{VQA}^{\text{T}}}, the relatively higher complexity of SQA triggers more frequent RVIS, which our adaptive framework effectively addresses by responding to RVIS.
% Pruning methods integrated with \proposed{} consistently outperform static baselines, with performance gains being particularly significant on SQA. 
% We attribute this to the relatively higher complexity of SQA compared to GQA and TextVQA, which leads to a more frequent occurrence of RVIS. As a result, our adaptive framework can respond to these dynamic visual shifts more effectively than static pruning approaches.

\subsection{In-Depth Analysis} \label{sec:indepth_analysis}

\begin{wraptable}{r}{0.50\textwidth} %
    \centering
\vspace{-8ex}
\caption{Ablation for RISD and CPTS.}
\setlength{\tabcolsep}{2.5pt}
\resizebox{1.0\linewidth}{!}{
        \begin{tabular}{lccccccc}
        \toprule
        \multirow{2.5}{*}{\textbf{Row}} & \multirow{2.5}{*}{\textbf{Detect}} & \multicolumn{2}{c}{\textbf{Swap}} & \multicolumn{2}{c}{\textbf{Qwen3 VL}} & \multicolumn{2}{c}{\textbf{InternVL3.5}} \\ \cmidrule(lr){3-4} \cmidrule(lr){5-6} \cmidrule(lr){7-8}
        & & \textbf{Strategy} & \textbf{Ratio\mm{\downarrow}} & \textbf{MV} & \textbf{MP} & \textbf{MV} & \textbf{MP} \\ \midrule
        (a) & \multicolumn{3}{c}{Full visual tokens} & 61.29 & 37.63 & 42.00 & 39.42 \\
        (b) & \multicolumn{3}{c}{FastV (33.3\%)} & 32.23 & 19.41 & 21.57 & 18.96 \\ \midrule
        (c) & Random & CPTS & 38.1\% & 37.69 & 22.36 & 31.09 & 22.18 \\
        (d) & Avg & CPTS & 38.1\% & 51.51 & 27.68 & 39.08 & 32.77 \\ 
        (e) & RISD & Full & 100\% & \textbf{53.04} & 29.55 & \textbf{40.96} & 34.16 \\
        (f) & RISD & Hard & 33.3\% & 45.05 & 27.26 & 36.12 & 30.69 \\
        (g) & RISD & Merge & 33.3\% & 47.58 & 28.38 & 37.16 & 32.54 \\ \midrule
        \rowcolor[rgb]{1, 0.9, 0.8} (h) & \textbf{RISD} & \textbf{CPTS} & {38.1\%} & 52.54 & \textbf{30.52} & 40.22 & \textbf{34.27} \\ \bottomrule
        \end{tabular}
}
\label{tab:ablation_study}
\vspace{-5.0ex}
\end{wraptable}
\textbf{Effect of Components.}
In~\cref{tab:ablation_study}, we evaluate the effectiveness of the RISD and CPTS modules against several baselines, using MathVerse (MV) and MMMU-Pro (MP).
\textit{\textbf{Effect of RISD:}}
We compare our fixed-threshold approach against two alternative detection strategies: \textit{Random} (row (c)), which triggers token swapping randomly with the same frequency as RISD, and 
\textcolor{black}{\textit{Avg} (row (d)), which utilizes a dynamic thresholding mechanism. Specifically, it calculates the running average of all preceding similarity scores and triggers CPTS when the current $s_l$ falls below this cumulative mean.}
Our RISD outperforms both alternatives, demonstrating that a constant threshold is sufficiently effective for capturing RVIS compared to either arbitrary timing or more complex running averaging.
\textbf{\textit{Effect of CPTS:}}
We evaluate various swap strategies during context-preserving generation. \textit{Hard} selection (row (f)), which discards previous visual tokens; the \textit{Merge} strategy (row (g)), \textcolor{black}{which compresses previous visual tokens into a single token and concatenates it with the newly identified visual tokens}; and \textit{Full} swap (row (e)), which serves as an upper bound that allows all visual tokens. \textcolor{black}{We also report the ratio of visual tokens during the context-preserving duration.}\footnote{\scriptsize For CPTS, ratio is reported as the average values, reflecting its dynamic token union mechanism.}
CPTS effectively matches the performance of the Full swap with only a marginal increase in the average token ratio. In contrast, Hard and Merge strategies exhibit significantly lower accuracy, highlighting their failure to adequately conserve the visual context required for precise generation.

\begin{wrapfigure}{r}{0.30\textwidth}
  \centering
  \vspace{-25pt}
  \includegraphics[width=\linewidth]{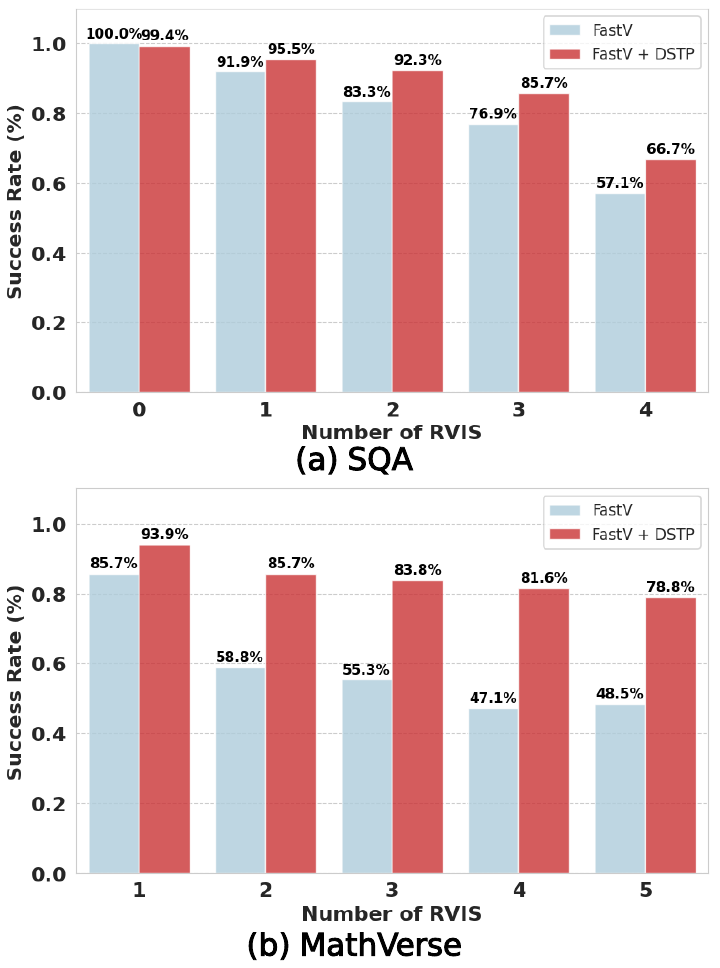}
  \vspace{-5ex}
  \caption{Success rate of FastV and \proposed{} across different RVIS frequencies.}
  \label{fig:motiv_improve}
  \vspace{-22pt}
\end{wrapfigure}
\smallskip \noindent 
\textbf{Robustness to RVIS.}
To validate the effectiveness of \proposed{}, we analyze success rates across varying RVIS frequencies (\cref{fig:motiv_improve}) while maintaining consistent settings (\cref{fig:rvis_causal}). We observe the following:
\textbf{1)} In both SQA and MathVerse, the performance of the baseline FastV~\cite{FastV} degrades significantly as the frequency of RVIS increases.
This decline underscores the inherent fragility of static pruning methods, which fail to adapt when encountering RVIS during decoding.
\textbf{2)} The application of \proposed{} yields consistent performance gains across all RVIS frequencies. Notably the performance gap between \proposed{} and FastV widens as RVIS becomes more frequent.
This widening margin validates the robustness of our framework in addressing RVIS and ensuring the model maintains focus on critical visual tokens during complex reasoning tasks.

\begin{wrapfigure}{r}{0.42\textwidth}
  \centering
  \vspace{-6ex}
  \includegraphics[width=0.95\linewidth]{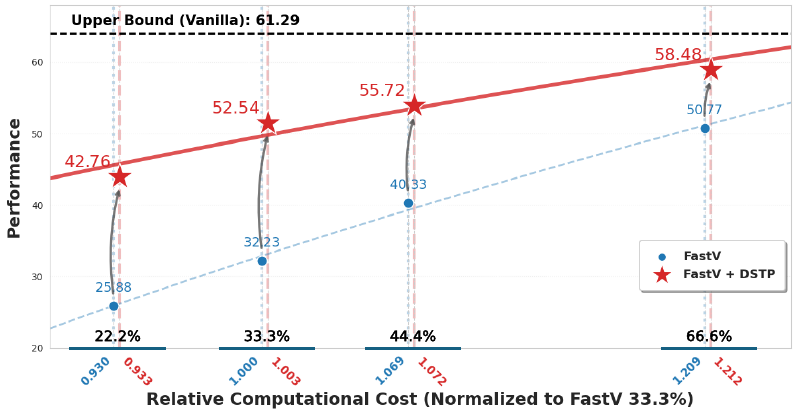}
  \caption{\textcolor{black}{Performance comparison of FastV versus \proposed{} across various computational cost on MathVerse~\cite{mathverse}}}
  \label{fig:various_ratio}
  \vspace{-26pt}
\end{wrapfigure}

\begin{figure}[t]
    \centering
    \includegraphics[width=1.0\linewidth]{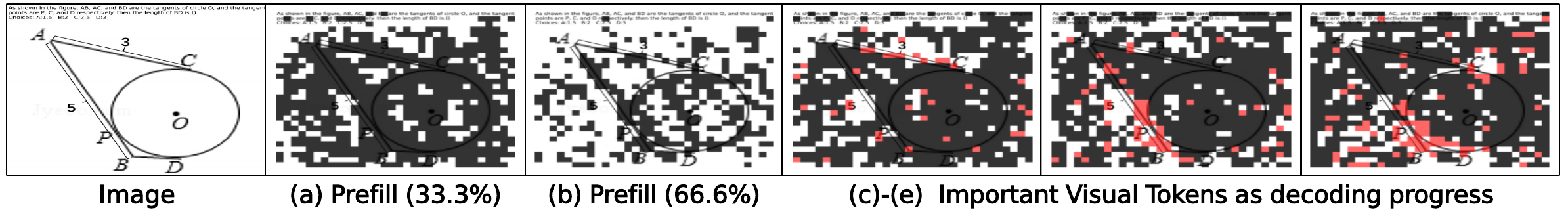}
    \caption{\small Visualization of visual token selection. (a) and (b) show FastV results at 33.3\% and 66.6\% token retention ratios. (c)-(e) illustrate \proposed{} results at a 33.3\% token retention ratio. White and black tokens represent retained and pruned tokens respectively. Red tokens denote those retained by \proposed{} but pruned even in the 66.6\% vanilla FastV.}
    \label{fig:showcase}
\end{figure}

\smallskip \noindent 
\textcolor{black}{\textbf{Comparative Analysis across TFLOPs.}
To ensure a rigorous evaluation,
we compare its performance against vanilla FastV across varying computational costs.
For this, we calculate TFLOPs, which accounts for the total floating-point operations executed during both prefill and decoding stages. For a consistent analysis, all values are normalized to the TFLOPs of vanilla FastV at a 33.3\% retention ratio (defined as 1.0).}\footnote{\label{note:TFLOps}\scriptsize Refer to the~\cref{app:K_TFLOPs} for detailed calculation of TFLOPs.}
\textcolor{black}{Our observations from~\cref{fig:various_ratio} are as follows:
\textbf{1)}
The additional computational cost incurred by \proposed{} is negligible across all retention ratios. Despite this closely matched cost, \proposed{} provides a substantial performance gains, suggesting that allocating resources adaptively is far more effective than static pruning.
\textbf{2)} 
Remarkably, \proposed{} at a 33.3\% ratio surpasses the performance of vanilla FastV at a much higher 66.6\% ratio, while requiring significantly less computational cost.
This highlights that providing relevant visual information at the precise decoding steps is far more critical than simply increasing the total number of tokens during the prefill stage.
Furthermore, as shown in~\cref{fig:showcase}, \proposed{} successfully retrieves essential visual context that remains pruned even by the 66.6\% static baseline, underscoring the superior adaptability of our framework.}

\begin{wrapfigure}{r}{0.33\textwidth}
  \centering
  \vspace{-24pt}
  \includegraphics[width=\linewidth]{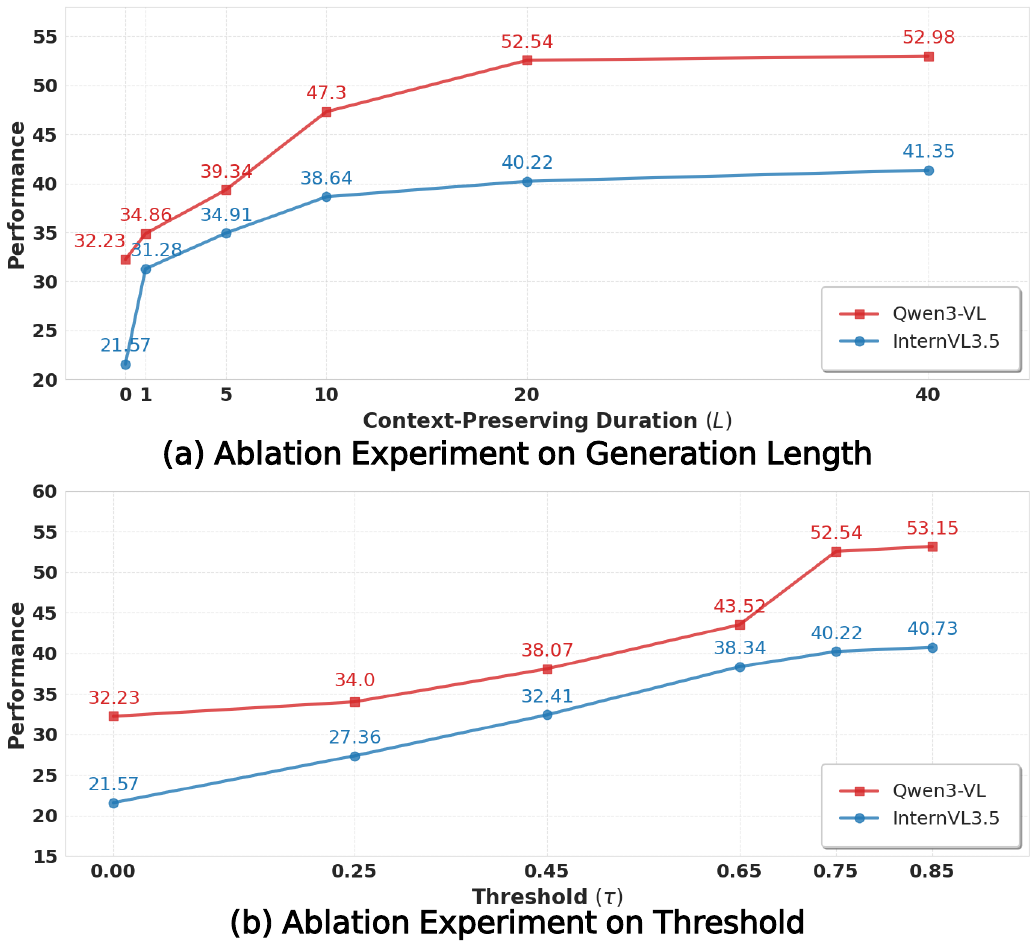}
  \vspace{-4ex}
  \caption{Hyper-parameter experiments on Generation Length \mm{L} and threshold \mm{\tau}.}
  \label{fig:hyperparameters}
  \vspace{-18pt}
\end{wrapfigure}

\smallskip \noindent
\textbf{Hyper-parameter Experiments.}
We evaluate the sensitivity of $L$ and $\tau$ on MathVerse in~\cref{fig:hyperparameters}, noting that $L=0$ and $\tau=0$ represent the vanilla FastV baseline. Our observations are as follows:
\textbf{1)} 
Performance scales with $L$, though a sharp decline occurs at lower values. 
This indicates that a sufficient duration
is essential for maintaining logical continuity and stable answer generation. Conversely, performance gains saturate as $L$ increases further, suggesting that a moderate duration is enough to ensure reasoning accuracy without unnecessary overhead.
\textbf{2)} 
\textcolor{black}{The threshold $\tau$ regulate the tradeoff between RVIS detection with computational cost.} At lower values, the model may fail to capture RVIS, resulting in reduced performance. While a higher $\tau$ enables more proactive context refocusing, excessively high values lead to computational inefficiency by triggering visual token swapping too frequently. Therefore, a moderate threshold is most effective for providing necessary visual evidence while maintaining operational efficiency.

% \section{Additional Experiment(?)}
\vspace{-2.5ex}
\subsection{Efficiency Analysis}
\vspace{-0.4ex}
While many pruning methods~\cite{SparseVLM, FastV, DivPrune} prioritize TFLOPs, which serves as a metric for theoretical computational complexity, recent research~\cite{deepspeed_laterncy_ref3, decoding_memory_bound_laterncy_ref4} emphasizes that latency is a more critical measure for realistic deployment. Following~\cite{dycoke_laterncy_ref2, aircache_laterncy_ref1}, we evaluate the computational efficiency of \proposed{} compared to the Vanilla and FastV models in~\cref{tab:efficiency}.\textsuperscript{\ref{note:TFLOps}}
% \footnote{\scriptsize Regarding the TFLOPs comparison, please refer to the~\cref{app:K_TFLOPs}}
Our observations are as follows:
\textbf{1)}
Considering TPS as a direct metric for inference speed, \proposed{} introduces a minor overhead compared to the FastV baseline due to the dynamic nature of its detect-and-swap mechanism during decoding. However, it still achieves significantly higher efficiency than the Vanilla model while maintaining superior accuracy. This demonstrates that \proposed{} is an effective framework that maximizes reasoning performance while keeping computational costs substantially lower than the vanilla model.
\textbf{2)}
Notably, \proposed{} achieves the lowest latency per example (Lat./Ex.) on Qwen3-VL, despite having a lower TPS than FastV. We attribute this efficiency gain to a reduction in the Average Tokens generated. Compared to FastV, \proposed{} produces fewer tokens while maintaining higher accuracy. This suggests that by providing the most relevant visual context at the appropriate reasoning steps, our framework prevents the model from engaging in redundant or unnecessary reasoning, thereby streamlining the generation process and reducing end-to-end inference time.
In \cref{app:L_Decoding_token_budget}, we provide additional experiments demonstrating the potential of \proposed{} to further enhance computational efficiency.

% Add-on Ration 설명 추가
% For further efficiency of \proposed{}, the decoding budget \mm{k} can 

\renewcommand{\arraystretch}{0.85}
\begin{table}[t]
\centering
\caption{\small Efficiency analysis of \proposed{} on FastV across Qwen3-VL and InternVL3.5. 
% Metrics include total latency, GPU memory, latency per example (Lat./Ex.), tokens per second (TPS), and average generation tokens.
}
\setlength{\tabcolsep}{2.5pt}
\label{tab:efficiency}
\resizebox{0.9\linewidth}{!}{
\begin{tabular}{lcccccc}
\toprule
\textbf{Method} & \textbf{Total Latency} $\downarrow$ & \textbf{GPU Mem.} $\downarrow$ & \textbf{Accuracy} $\uparrow$ & \textbf{Lat./Ex.} $\downarrow$ & \textbf{TPS} $\uparrow$ & \textbf{Avg Tokens} $\downarrow$ \\ \midrule

\rowcolor[gray]{.95} \multicolumn{7}{c}{Qwen3-VL-4B} \\ \midrule
Vanilla & 27:51:28 & 18.6G & 61.29 & 127.27s & 19.30 & 2419.4 \\
FastV & 26:07:27 & 14.6G & 32.23 & 119.35s & \textbf{24.99} & 2931.7 \\
\rowcolor[rgb]{1, 0.9, 0.8} FastV + \proposed{} & \textbf{22:26:49} & 15.1G & 52.54 & \textbf{102.55s} & 24.11 & \textbf{2412.7} \\ \midrule

\rowcolor[gray]{.95} \multicolumn{7}{c}{InternVL-3.5-8B} \\ \midrule
Vanilla & 2:34:53 & 28.9G & 42.00 & 11.79s & 31.55 & 377.3 \\
FastV & \textbf{2:00:03} & 25.2G & 21.57 & \textbf{9.14s} & \textbf{34.46} & 296.8 \\
\rowcolor[rgb]{1, 0.9, 0.8} FastV + \proposed{} & 2:11:12 & 26.3G & 40.22 & 9.99s & 33.79 & \textbf{227.8} \\ \bottomrule
\end{tabular}
}
\end{table}

\section{Related Works}
% \subsection{Visual Token Pruning for MLLMs}
% We report more detailed related works in \cref{app:M_Detailed_Related_Works}.
We report a more comprehensive discussion of related works in~\cref{app:M_Detailed_Related_Works}.

\smallskip \noindent
\textbf{Visual Token Pruning for MLLMs.} To mitigate the computational bottlenecks of MLLMs, visual token pruning has evolved into two main streams: training-based and training-free. Training-based methods (e.g., LLaVolta~\cite{llavolta}, ZipR1~\cite{ZipR1}, VCM~\cite{vcm}) incorporate learnable modules to enforce sparsity but incur significant overhead. Consequently, training-free methods have gained prevalence, utilizing three primary strategies: (1) Vision-Encoder Based: LLaVA-PruMerge~\cite{llava-prumerge} and VisionZip~\cite{VisionZip} utilize encoder-side features like [CLS] attention or CLIP-generated scores. (2) LLM-Based: FastV~\cite{FastV}, ZipVL~\cite{ZipVL}, and PDrop~\cite{PDrop} prune based on internal LLM attention, while DivPrune~\cite{DivPrune} treats pruning as a diversity maximization problem. (3) Cross-Modal: SparseVLM~\cite{SparseVLM} and SparseVILA~\cite{sparsevila} leverage query-aware interactions to guide sparsification, with the latter attempting token retrieval during decoding.
% \footnote{For more detailed related works, please refer to~\cref{app:M_Detailed_Related_Works}}

\smallskip \noindent \textbf{Evaluation of Visual Token Pruning.} Recent benchmarks like LLMC+~\cite{VTP_various_task_3} and UniPruneBench~\cite{VTP_various_task_2} have critically re-examined token pruning metrics, revealing that high pruning ratios often cause severe degradation in detail-sensitive tasks. While existing methods successfully accelerate inference, they frequently discard crucial visual cues required for deep understanding. Our research specifically targets complex visual reasoning, aiming to preserve and effectively retrieve critical information lost in static token pruning, thereby achieving robust reasoning performance without sacrificing efficiency.

% \subsection{Visual Reasoning in MLLMs}

% As MLLMs advance, evaluation has shifted from end-to-end accuracy to assessing true visual comprehension. Recent benchmarks challenge models to interpret visual inputs without relying on textual shortcuts; for instance, MathVerse~\cite{mathverse} and MMMU-Pro~\cite{mmmu-pro} isolate visual cues to test genuine diagram interpretation. Furthermore, DynaMath~\cite{dynamath} reveals that models often lack robustness, struggling with minor visual variations in mathematical problems.

% Beyond outcome-based metrics, researchers are increasingly scrutinizing the underlying reasoning process. We-Math~\cite{wemath} and MathVision~\cite{mathvision} demonstrate that MLLMs frequently rely on rote memorization rather than human-like generalization, while LogicVista~\cite{logicvista} emphasizes the necessity of logical cognition for spatial navigation and puzzles.

% Collectively, these studies underscore that complex visual reasoning demands precise and uncompromised visual details. This motivates our work: while conventional pruning methods inadvertently discard the very information required for these benchmarks, our approach is designed to preserve and retrieve essential visual cues, ensuring robust performance in advanced reasoning scenarios.

\vspace{-2ex}
\section{Conclusion}
\vspace{-1ex}
% In this work, we revisit a key assumption in existing visual token pruning methods: that the visual tokens deemed relevant at the prefill stage remain sufficient throughout decoding, even for complex visual reasoning. We argue that this assumption is often violated by Relevant Visual Information Shift (RVIS), where the model’s visual focus changes over decoding steps.
% To address RVIS, we propose \proposed{}, a training-free, plug-and-play add-on framework that adaptively realigns visual tokens with the model’s shifting attention during decoding. This simple mechanism substantially mitigates performance degradation while incurring minimal overhead.

In this paper, we identify Relevant Visual Information Shift (RVIS) as a critical failure driver in static pruning methods during complex reasoning. To address this, we propose \proposed{}, a training-free, add-on framework that adaptively aligns visual tokens with the model's shifting focus throughout the decoding stage, significantly mitigating performance degradation with minimal overhead.
% In this paper, we identify Relevant Visual Information Shift (RVIS) as a primary failure driver that fundamentally limits the efficacy of existing static visual token pruning methods, particularly in reasoning-heavy contexts. 
% Our analysis reveals that unlike simple visual understanding, visual reasoning requires a dynamic re-focus of visual tokens to meet shifting informational needs throughout the decoding process.
% To address this, we propose \textbf{D}ecoding-stage \textbf{S}hift-aware \textbf{T}oken \textbf{P}runing (\proposed{}), a training-free add-on framework that empowers previous pruning methods to adaptively align visual tokens with the model's shifting focus during the decoding stage.
% Extensive experiments demonstrate that \proposed{} significantly mitigates performance degradation in complex reasoning tasks with minimal computational overhead.

\smallskip \noindent 
% We believe this work provides a novel perspective by demonstrating that visual token pruning must account for RVIS during the decoding stage, moving beyond the traditional focus on the initial prefill stage.
% Overall, we believe this work offers a new perspective: visual token pruning should accounts for RVIS. Our findings underscore that addressing these decoding-stage dynamics is essential for ensuring reasoning integrity while sustaining the efficiency benefits of pruning.
We believe this work provides a novel perspective by demonstrating that visual token pruning must move beyond the prefill stage to account for RVIS during decoding.
Our findings underscore that addressing these decoding-stage dynamics is essential for ensuring reasoning integrity while sustaining the efficiency benefits of pruning.

\bibliographystyle{splncs04}
\bibliography{main}

\clearpage

\appendix
\renewcommand{\thefigure}{\Roman{figure}}
\renewcommand{\thetable}{\Roman{table}}
\renewcommand{\thealgorithm}{\Roman{algorithm}}

\setcounter{figure}{0}
\setcounter{table}{0}
\newpage
\startcontents[appendix]

\section{Differences between Visual Understanding and Visual Reasoning abilities}
\label{app:A_task_diff}
\vspace{-1ex}
MLLM evaluation broadly distinguishes two capability axes: \textbf{Visual Understanding} and \textbf{Visual Reasoning}. The former concerns recognizing \emph{what} appears in an image, while the latter requires multi-step logical inference about \emph{how} and \emph{why} visual elements relate.

\smallskip \noindent
\textbf{Visual Understanding.}
Visual understanding refers to the ability to identify objects, scenes, and basic attributes within an image. Given a question such as \textit{What is in this image?}, the model matches visual patterns—textures, colors, shapes—to known labels or concepts. Standard benchmarks~\cite{TextVQA, GQA, ScienceQA} assess this through direct, fact-based questions about the presence or simple properties of objects. A defining characteristic of these tasks is that the relevant visual cues are typically concentrated in a narrow image region, which can be identified during the prefill stage and remains sufficient throughout generation.

\smallskip \noindent
\textbf{Visual Reasoning.}
Visual reasoning, by contrast, demands higher-order logical thinking: interpreting spatial relationships, performing multi-step derivations, and integrating information from multiple image regions. Benchmarks such as MathVerse~\cite{mathverse}, MathVision~\cite{mathvision}, WeMath~\cite{wemath}, DynaMath~\cite{dynamath}, MMMU-Pro~\cite{mmmu-pro}, and LogicVista~\cite{logicvista} evaluate this ability by requiring rigorous mathematical or logical derivations that cannot be solved through simple pattern matching alone. As identified in Sec. 3.2, a key characteristic of reasoning tasks is \textit{Relevant Visual Information Shift (RVIS)}: the model must dynamically transition its visual focus across different image regions as each successive reasoning step introduces new informational requirements.

\vspace{-1ex}
\section{Additional Attention Heatmap Visualizations for VQA and VMR}
\label{app:B_vqa_heatmap}
\vspace{-1ex}

\begin{figure*}[t]
    \centering
    \includegraphics[width=0.8\linewidth]{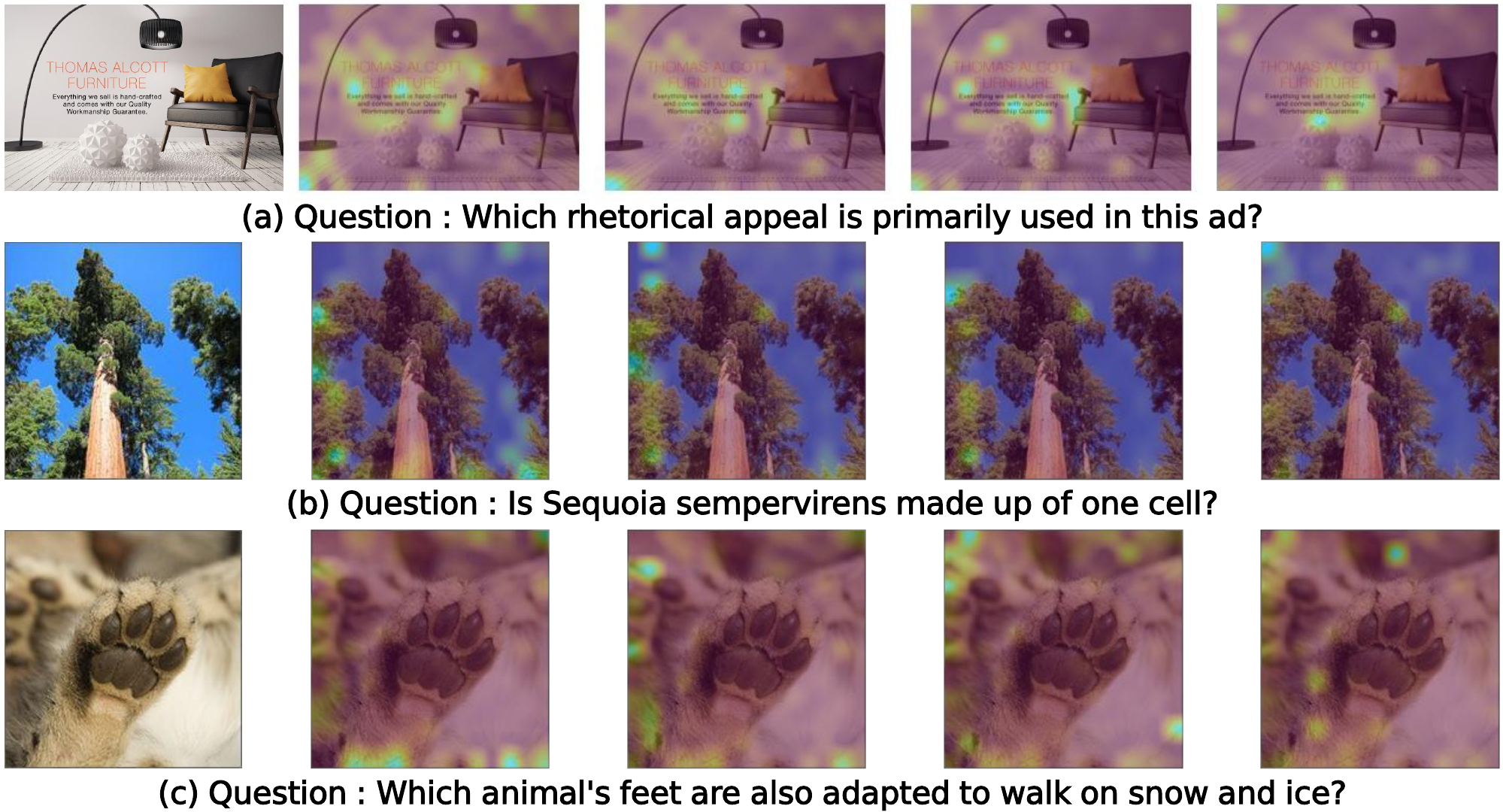}
    \vspace{-2ex}
    \caption{\footnotesize Attention heatmaps for ScienceQA (VQA) samples. Decoding steps progress from left to right.}
    \label{fig:app_vqa_vis}
    \vspace{-3ex}
\end{figure*}

\begin{figure*}[t]
    \centering
    \includegraphics[width=0.8\linewidth]{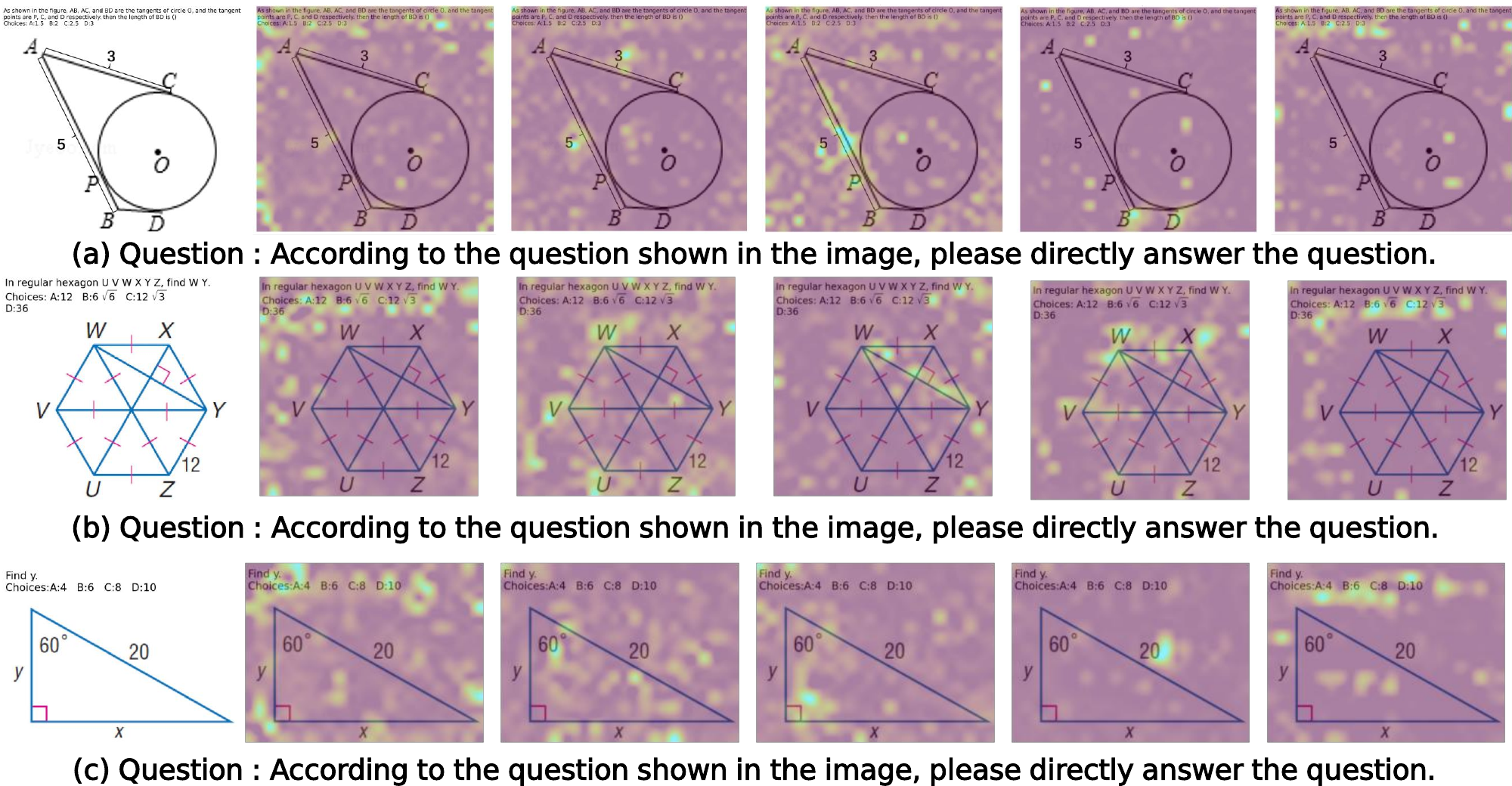}
    \vspace{-1ex}
    \caption{\footnotesize Attention heatmaps for MathVerse (VMR) samples. Decoding steps progress from left to right.}
    \label{fig:app_vmr_vis}
    \vspace{-2ex}
\end{figure*}

This section presents additional attention heatmaps that illustrate how Relevant Visual Information Shift (RVIS) manifests differently across VQA and VMR benchmarks.

\smallskip \noindent
As shown in~\cref{fig:app_vqa_vis}, VQA samples consistently exhibit no visual information shift throughout the decoding process, regardless of the input characteristics. 
In~\cref{fig:app_vqa_vis}(a), where the answer depends on a localized image region (\eg, the textual area), the model's attention remains firmly anchored to that region throughout the generation without shifting toward irrelevant objects such as chairs or lamps. In~\cref{fig:app_vqa_vis}(b) and (c), where the image itself has low relevance and the text query alone suffices, attention clusters in the background or at image corners, consistent with the \textit{Attention Sink} phenomenon reported in prior work. Crucially, in both cases, the model's visual focus remains static, confirming that RVIS does not arise in simple visual understanding tasks.

\smallskip \noindent
In stark contrast,~\cref{fig:app_vmr_vis} reveals that VMR samples exhibit repeated RVIS events as decoding progresses. The model's visual attention does not stay fixed on any single region; instead, it continuously shifts across distinct image areas at each reasoning step to gather the specific visual evidence required for the current logical derivation. This persistent shifting behavior is consistent with the findings in Fig. 1, reinforcing that RVIS is an inherent and recurring characteristic of multi-step visual reasoning.
%—fundamentally distinguishing it from the static attention patterns observed in VQA.

\vspace{-2ex}
\section{Detailed Algorithm for RVIS Detection}
\label{app:C_find_peak}
\vspace{-1ex}

\begin{wrapfigure}{r}{0.5\textwidth}
\begin{minipage}{\linewidth}
\vspace{-45pt}
\begin{algorithm}[H]
\caption{RVIS Detection Procedure}
\label{alg:find_rvis}
\begin{algorithmic}[1]
\renewcommand{\algorithmiccomment}[1]{\\ \quad {\scriptsize \color{gray} // #1}}
\renewcommand{\algorithmicensure}{\textbf{Return:}}
\REQUIRE $a_{t_0 \to v}$, $\{a_{t_l \to v}\}_{l=1}^{L}$, threshold $\tau$
\ENSURE Set of RVIS indices $P$

\STATE $S \leftarrow []$
\FOR{$l = 1$ \TO $L$} 
    \STATE $s_l \leftarrow \text{Sim}(a_{t_0 \to v}, a_{t_l \to v})$ % \COMMENT{Calculate similarity for each step}
    \STATE $S.\text{append}(s_l)$
\ENDFOR

\STATE $\mathcal{I}_{\text{cand}} = \{l \mid s_l < \tau\}$ % \COMMENT{Filter potential candidates below $\tau$}
\STATE $\mc{I}_{\text{peaks}}=\text{find\_local\_minima}(S)$
% \STATE $P = \mc{I}_{\text{peaks}} \cap \mathcal{I}_{\text{cand}}$ 
\RETURN $P = \mc{I}_{\text{peaks}} \cap \mathcal{I}_{\text{cand}}$
\end{algorithmic}
\end{algorithm}
\end{minipage}
\vspace{-5ex}
\end{wrapfigure}
\noindent
This section describes the algorithm used to detect RVIS in a systematic and reproducible manner, as summarized in~\cref{alg:find_rvis}. The detection relies on a two-stage filtering process applied to the cosine similarity \mm{s_l} between the prefill-stage attention \mm{a_{t_0 \to v}} and the attention \mm{a_{t_l \to v}} at each decoding step \mm{l}. 
In the first stage, we compute \mm{s_l} across all decoding steps and flag every step whose similarity falls below a predefined threshold \mm{\tau} as a candidate for an information shift. 
However, naively counting all such steps would significantly overestimate the number of RVIS events, since a single RVIS typically spans a contiguous sequence of low-similarity steps rather than appearing as an isolated point.
To address this, the second stage applies the \textit{scipy.signal.find\_peaks} algorithm to locate the local minima within the similarity signal, consolidating each contiguous low-similarity region into a single discrete RVIS event. The final set of detected RVIS indices is obtained by intersecting these local minima with the candidate set from the first stage.

\vspace{-1ex}
\section{Additional Experiments of RVIS Reasoning-Intrinsic Nature}
\label{app:D_RVIS_Nature_additional}
\vspace{-1ex}

This section presents additional experiments that further substantiate RVIS as an inherent property of the reasoning process, extending the analysis in Sec. 3.1.

\begin{wrapfigure}{r}{0.44\linewidth}
    \centering
    \vspace{-22pt}
    \includegraphics[width=0.99\linewidth]{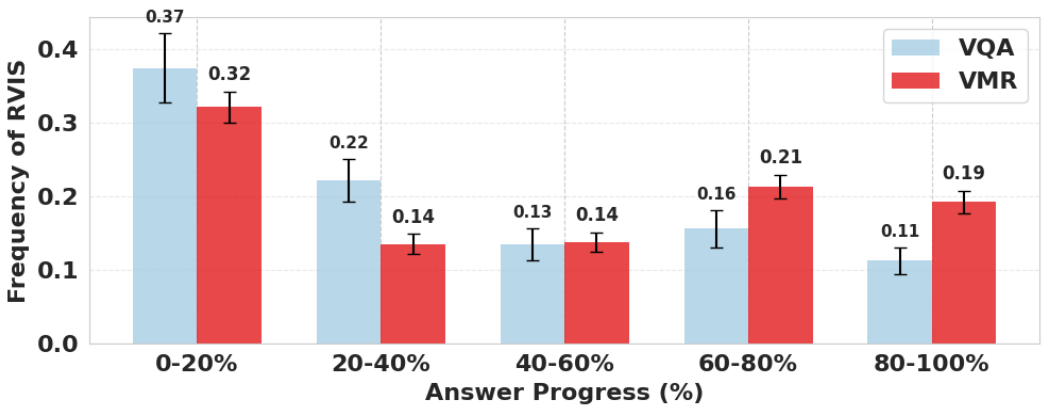}
    \vspace{-4ex}
    \caption{\small Temporal occurrence distribution of RVIS.
    % for ScienceQA (VQA) and MathVerse (VMR).
    }
    \vspace{-23pt}
    \label{fig:rvis_temporal}
\end{wrapfigure}

\smallskip \noindent
\textbf{Temporal Distribution across VQA and VMR.}
To examine \textit{when} RVIS occurs during generation, we analyze its temporal distribution across VQA and VMR. 
We collect samples that exhibit at least one RVIS event and measure shift frequency across normalized generation progress (0\% to 100\%), as shown in~\cref{fig:rvis_temporal}. 
In VQA, RVIS events are concentrated within the first 20\% of generation, whereas in VMR, shifts continue to occur well beyond the 60\% mark. 
This contrast indicates that the temporal occurrence of RVIS is governed by the reasoning demands of the task: unlike VQA, where an early visual focus suffices, multi-step reasoning requires sustained visual re-focus throughout the entire generation process.

\begin{wrapfigure}{r}{0.44\linewidth}
    \centering
    \vspace{-22pt}
    \includegraphics[width=0.99\linewidth]{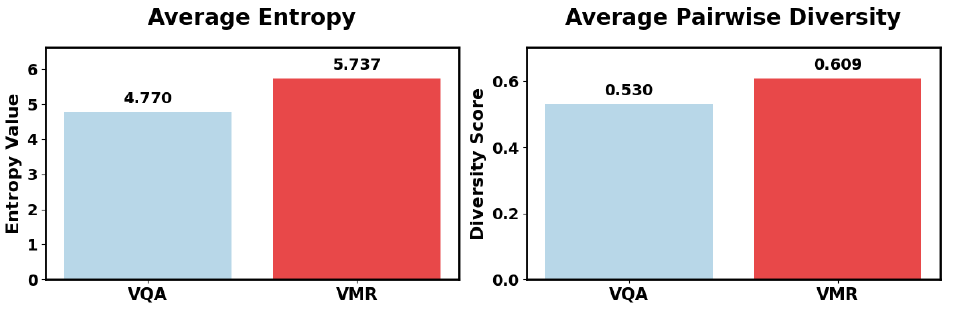}
    \vspace{-4ex}
    \caption{\small Temporal occurrence distribution of RVIS.
    % for ScienceQA (VQA) and MathVerse (VMR).
    }
    \vspace{-23pt}
    \label{fig:RVIS_Ent_Div}
\end{wrapfigure}

\smallskip \noindent
\textbf{Visual Focus Diversity.}
Beyond temporal patterns, we examine how distinct successive RVIS events are from one another using two complementary metrics, as shown in~\cref{fig:RVIS_Ent_Div}. Spatial Entropy measures how broadly each RVIS attention vector is spread across the image, distinguishing wide-range exploration from narrow fixation. 
Pairwise Diversity, defined as 1 - \text{similarity} between RVIS attention vectors, quantifies how different the visual focus is across distinct RVIS events.
VMR exhibits substantially higher values in both metrics compared to VQA, indicating that while VQA shifts remain narrowly localized, RVIS events in VMR explore broader regions and attend to distinct image areas across different reasoning steps.

\vspace{-1ex}
\section{Detailed Algorithm of \proposed{}}
\label{app:E_method_algorithm}
\vspace{-1ex}
We provide the detailed procedure of \proposed{} in~\cref{alg:proposed_method_detailed}, where $\bm{X^{\text{Active}}_v}$ denotes the set of visual tokens currently accessible to the MLLM.
\begin{algorithm}[H]
\caption{Detailed Operational Flow of \proposed{}}\label{alg:proposed_method_detailed}
\resizebox{0.85\linewidth}{!}{%
\begin{minipage}{\linewidth}
\begin{algorithmic}[1]
\renewcommand{\algorithmiccomment}[1]{\hfill {\scriptsize \color{gray} // #1}}
\REQUIRE Visual tokens $\bm{X_v}$, text query tokens $\bm{X_t}$, detection threshold $\bm{\tau}$, context-preserving duration $\bm{L}$, token retention ratio $\bm{R}$
\STATE \textbf{Return:} Generated response sequence $\bm{Y}$
\STATE
\STATE \textit{\textbf{Prefill Stage}}
\STATE Select top-$K$ tokens $\bm{X^{\text{Pruned}}_v}$ using base pruning methods, where $K = R \times |\bm{X_v}|$
\STATE Reserve discarded tokens $\bm{X^{\text{Reserved}}_v} \leftarrow \bm{X_v} - \bm{X^{\text{Pruned}}_v}$
\STATE $\bm{X^{\text{Active}}_v} \leftarrow \bm{X^{\text{Pruned}}_v}$
\STATE
\STATE \textit{\textbf{Decoding Stage}}
\STATE $\text{REMAIN\_CP} \leftarrow 0$ \COMMENT{Initialize context-preserving counter}
\WHILE{Generation continues}
    \IF{$\text{REMAIN\_CP} == 0$}
        \STATE \textbf{\textit{RISD Module}}
        \STATE Compute attention $\bm{a^{\text{Pruned}}_{t_l \to v}}$ using Eq.~(4)
        \STATE Calculate similarity $s_l$ using Eq.~(5)
        \IF{$s_l < \tau$}
            \STATE \textbf{\textit{CPTS Module}} \COMMENT{RVIS detected, invoke CPTS}
            \STATE Calculate attention $\bm{a^{\text{Reserved}}_{t_l \to v}}$ using Eq.~(4)
            \STATE Reconstruct full attention $\bm{a_{t_l \to v}}$ using Eq.~(6)
            \STATE Re-select top-$K$ tokens $\bm{X^{\text{Pruned}, l}_v}$ based on $\bm{a_{t_l \to v}}$
            \STATE Extract $\bm{X^{CP, l}_v}$ using Eq.~(7)
            \STATE $\bm{X^{\text{Active}}_v} \leftarrow \bm{X^{CP, l}_v}$
            \STATE $\text{REMAIN\_CP} \leftarrow L$
        \ENDIF
    \ELSE
        \STATE $\text{REMAIN\_CP} \leftarrow \text{REMAIN\_CP} - 1$
        \IF{$\text{REMAIN\_CP} == 0$}
            \STATE $\bm{X^{\text{Active}}_v} \leftarrow \bm{X^{\text{Pruned}}_v}$
        \ENDIF
    \ENDIF
\ENDWHILE
\STATE \textbf{return} $\bm{Y}$
\end{algorithmic}
\end{minipage}}
\end{algorithm}
\vspace{-6ex}

\vspace{-1ex}
\section{Additional Experiments on Token Reversion}
\label{app:F_Token_Reversion}
\vspace{-1ex}

\begin{wraptable}{r}{0.43\textwidth}
\centering
\vspace{-7.5ex}
\caption{Ablation for Reversion Strategy on FastV.}
\setlength{\tabcolsep}{2.5pt}
\resizebox{1.0\linewidth}{!}{
    \begin{tabular}{lcc}
    \toprule
    \multirow{2.5}{*}{\textbf{Method}} & \multicolumn{2}{c}{\textbf{Qwen3-VL}} \\ \cmidrule(lr){2-3}
    & \textbf{MathVerse} & \textbf{MMMU-Pro} \\ \midrule
    Vanilla                   & 61.29 & 37.63 \\
    FastV (33.3\%)            & 32.23 & 19.41 \\ \midrule
    \proposed{} w/o Reversion & 44.04 & 22.03 \\
    \rowcolor[rgb]{1, 0.9, 0.8} \textbf{\proposed{}} & \textbf{52.54} & \textbf{30.52} \\ \bottomrule
    \end{tabular}
}
\label{tab:reversion_ablation}
\vspace{-3.5ex}
\end{wraptable}
This section examines the reversion strategy introduced in Sec. 4.2, corresponding to Line~20 of~\cref{alg:proposed_method_detailed}.
In the standard \proposed{} protocol, the model reverts to the initial prefill-stage set $X^{\text{Pruned}}_v$ once the context-preserving duration $L$ expires. Here, we evaluate an alternative: instead of reverting, we re-select a new token set $X^{\text{Pruned}, l+L}_v$ based on the query $q_{l+L}$ at the step where the duration ends. As shown in~\cref{tab:reversion_ablation}, this re-selection variant does not improve performance and even degrades it.
We attribute this to the nature of the tokens selected at each stage. As shown in Fig. 1 and~\cref{fig:app_vmr_vis}, the newly identified tokens $X^{\text{Pruned}, l}_v$ during RVIS capture fine-grained, step-specific details for a particular reasoning step, whereas the prefill-stage tokens $X^{\text{Pruned}}_v$ encode the broader global context essential for coherent generation. Re-selecting at step $l+L$ based on a localized reasoning state risks discarding this global context without compensating reasoning benefits. This validates our reversion strategy: reverting to $X^{\text{Pruned}}_v$ after the context-preserving duration effectively balances step-specific visual retrieval with global image understanding.

\vspace{-2ex}
\section{More Generalization Experiment}
\label{app:G_General_Experiment}
\vspace{-1ex}
\begin{table*}[t]
\centering
\caption{Generalization performance of \proposed{} on Thinking-based MLLMs and Large-scale MLLMs using MathVerse and MMMU-Pro benchmarks. All experiments are conducted with a token retention ratio of 33.3\%.}
\vspace{-1ex}
\setlength{\tabcolsep}{5pt}
\resizebox{0.95\linewidth}{!}{
    \begin{tabular}{lcc|lcc}
    \toprule
    \multicolumn{3}{c}{\textbf{Thinking MLLMs}} & \multicolumn{3}{c}{\textbf{Large-scale MLLMs}} \\
    \cmidrule(lr){1-3} \cmidrule(lr){4-6}
    \textbf{Method} & \textbf{MathVerse} & \textbf{MMMU-Pro} & \textbf{Method} & \textbf{MathVerse} & \textbf{MMMU-Pro} \\ \midrule
    \rowcolor[gray]{0.95} \multicolumn{3}{c}{\textbf{Qwen3-VL-4B-Thinking}} & \multicolumn{3}{c}{\textbf{Qwen3-VL-8B}} \\
    Vanilla (Full Tokens) & 73.01 & 40.51 & Vanilla (Full Tokens) & 62.81 & 42.31 \\
    FastV (33.3\%) & 40.28 & 29.77 & FastV (33.3\%) & 37.64 & 18.64 \\
    w/ \proposed{} & 58.49 & 36.52 & w/ \proposed{} & 51.16 & 33.31 \\ \midrule
    \rowcolor[gray]{0.95} \multicolumn{3}{c}{\textbf{InternVL3.5-8B-Thinking}} & \multicolumn{3}{c}{\textbf{InternVL3.5-14B}} \\
    Vanilla (Full Tokens) & 47.46 & 37.68 & Vanilla (Full Tokens) & 47.89 & 44.39 \\
    FastV (33.3\%) & 18.48 & 20.63 & FastV (33.3\%) & 31.86 & 31.66 \\
    w/ \proposed{} & 37.25 & 31.40 & w/ \proposed{} & 43.95 & 40.64 \\
    \bottomrule
    \end{tabular}
}
\label{tab:app_generalization_fixed}
\vspace{-2ex}
\end{table*}

To validate the backbone-agnostic nature of \proposed{}, we evaluate it on two additional MLLM categories: (i) \textbf{Thinking-based MLLMs} (Qwen3-VL-4B-Thinking, InternVL3.5-8B-Thinking), which employ a think-and-answer paradigm for deeper reasoning, and (ii) \textbf{Large-scale MLLMs} (Qwen3-VL-8B, InternVL3.5-14B), which test scalability. As shown in~\cref{tab:app_generalization_fixed}, FastV suffers significant degradation across all variants, while integrating \proposed{} consistently leads to substantial recovery. This confirms that \proposed{} is robust to both model scale and answering paradigm, generalizing effectively from standard MLLMs to their thinking-based and larger-scale counterparts.

\vspace{-1.5ex}
\section{Additional Implementation Details}
\label{app:H_Implementation_Detail}
\vspace{-1.5ex}
This section provides implementation details regarding the experimental environment, pruning configurations, and model-specific settings.

\smallskip \noindent
\textbf{Evaluation and Framework and Hardware.}
All evaluations were conducted using the \textit{VLMEvalKit} framework to ensure consistent prompt templates and scoring metrics across models.
Experiments were run on 4 NVIDIA L40S GPUs, with the maximum generation length set to 4,096 tokens to avoid truncating complex reasoning chains. Efficiency-related metrics (latency, GPU memory, tokens per second) were measured on a single L40S GPU.

\smallskip \noindent
\textbf{Pruning Methods Configuration.}
All baseline pruning methods were configured following the hyperparameters reported in their respective papers. FastV~\cite{FastV} initiates pruning at the second layer ($L=2$) of the LLM backbone. DivPrune~\cite{DivPrune} is applied directly to the vision encoder output. VisionZip~\cite{VisionZip} extracts features from the penultimate layer of the vision encoder.

\smallskip \noindent
\textbf{Model Generation Configurations.}
All evaluations use deterministic greedy decoding with \texttt{do\_sample=False}, \texttt{temperature=1.0}, \texttt{top\_p=1.0}, \texttt{num\_beams=1}, and \texttt{max\_new\_tokens=4096}. For image processing, Qwen3-VL constrains resolutions between $256 \times 28^2$ and $1{,}280 \times 28^2$ total pixels. InternVL3.5 uses the default dynamic image size setting with a maximum of 12 sub-blocks.

\vspace{-1ex}
\section{Comprehensive Analysis for visual reasoning benchmarks}
\label{app:I_intensive_analysis}
\vspace{-1ex}
This section investigates how task difficulty and visual dependency impact pruning effectiveness. We categorize benchmarks into two groups based on their \textbf{visual intensity} to analyze the performance gains of \proposed{}.

\smallskip \noindent
\textbf{Visual-Intensive Benchmarks (G1).}
Benchmarks such as MMMU-Pro~\cite{mmmu-pro} and MathVerse~\cite{mathverse} are classified as visual-intensive because they embed critical information directly within images to prevent \textit{text-shortcuts}~\cite{ARES, mmmu-pro}. These benchmarks not only demand rigorous interpretation of complex diagrams and charts but also involve more challenging reasoning, requiring the model to perform multi-step logical derivations that cannot be solved through linguistic patterns alone. In such settings, static pruning methods are particularly vulnerable, as they permanently discard the fine-grained visual details needed to sustain complex reasoning from the outset.

\smallskip \noindent
\textbf{Visual-Lite Benchmarks (G2).}
In contrast, benchmarks like WeMath~\cite{wemath} and DynaMath~\cite{dynamath} exhibit a more balanced dependency between visual and textual modalities. They provide substantial reasoning cues within the text instructions, allowing the model to leverage textual context to support its reasoning. As a result, the negative impact of visual token pruning is relatively alleviated compared to G1.

\smallskip \noindent
\textbf{Comparative Analysis of Visual Intensity.}
As shown in~\cref{tab:group_improvement}, \proposed{} yields  
\begin{wraptable}{r}{0.45\textwidth}
\centering
\vspace{-32pt}
\caption{Average Performance Improvement (\%) of Proposed Method over Baselines (Retain 33.3\% Tokens).}
\label{tab:group_improvement}
\small
\setlength{\tabcolsep}{4pt}
\resizebox{1.0\linewidth}{!}{
\begin{tabular}{l|cc|cc}
\toprule
\multirow{2}{*}{\textbf{Method}} & \multicolumn{2}{c|}{\textbf{Qwen3-VL-4B}} & \multicolumn{2}{c}{\textbf{InternVL3.5-8B}} \\
& \textbf{G1} & \textbf{G2} & \textbf{G1} & \textbf{G2} \\ \midrule
FastV & 60.13 & 42.94 & 83.61 & 52.87 \\
DivPrune & 81.26 & 41.14 & 54.80 & 35.53 \\
VisionZip & 52.68 & 26.04 & 60.16 & 41.73 \\ \midrule
\textbf{Total Avg.} & \textbf{64.69} & \textbf{36.71} & \textbf{66.19} & \textbf{43.38} \\ \bottomrule
\end{tabular}
}
\vspace{-17pt}
\end{wraptable}
significantly higher performance gains in G1 (Visual-Intensive) compared to G2 (Visual-Lite). Since G1 benchmarks combine heavy visual dependency with more demanding reasoning complexity, RVIS occurs more frequently and with greater impact, making \proposed{}'s adaptive token swapping particularly effective. 
These results confirm that our framework provides the largest benefits precisely where they are most needed: tasks requiring both deep visual understanding and complex multi-step reasoning.

\vspace{-2ex}
\section{Expanded Experiment Results on VQA}
\label{app:J_VQA}
\vspace{-2ex}

In~\cref{tab:app_visual_understanding}, we present expanded VQA results  corresponding to the subset reported in Tab. 2.

\renewcommand{\arraystretch}{0.95}
\begin{table*}[t]
\centering
\caption{\textbf{Visual Understanding} Performance of \textbf{Qwen3-VL-4B} and \textbf{InternVL3.5-8B} under different token retention ratios with various VTP methods and our method. Values in parentheses are the proportion relative to the upper bound.}
\label{tab:app_visual_understanding}
\vspace{-2ex}
\resizebox{\linewidth}{!}{
\begin{tabular}{c|ccc|c|ccc|c}
% \specialrule{0.1em}{0em}{0em} \\
\toprule
\textbf{Model} & \multicolumn{4}{c|}{\textbf{Qwen3-VL-4B~\cite{qwen3vl}}} & \multicolumn{4}{c}{\textbf{InternVL3.5-8B~\cite{internvl3_5}}} \\ \midrule
\textbf{Method} & \textbf{ScienceQA} & \textbf{TextVQA} & \textbf{GQA} & \textbf{Acc. (\%)} & \textbf{ScienceQA} & \textbf{TextVQA} & \textbf{GQA} & \textbf{Acc. (\%)} \\ \specialrule{0.1em}{0em}{0em}
\rowcolor{gray!20} \multicolumn{9}{c}{\textit{Upper Bound, Full Visual Tokens (100\%)}} \\
Vanilla & 93.42 & 81.57 & 61.82 & 100\% & 96.70 & 77.59 & 61.24 & 100\% \\
 \specialrule{0.1em}{0em}{0em}

%% Retain 33.3%
\rowcolor{gray!20} \multicolumn{9}{c}{\textit{Retain 33.3\% Tokens ($\downarrow$ 66.7\%)}} \\ 
FastV \scriptsize{(ECCV'24)} & 87.98 \scriptsize{(94.2\%)} & 74.82 \scriptsize{(91.7\%)} & 60.04 \scriptsize{(97.1\%)} & 94.3\% & 81.60 \scriptsize{(84.4\%)} & 64.92 \scriptsize{(83.7\%)}& 57.08 \scriptsize{(93.2\%)}& 87.1\% \\
w/ \proposed{} & 91.84 \scriptsize{(98.3\%)} & 77.95 \scriptsize{(95.6\%)} & 60.93 \scriptsize{(98.6\%)} & 97.5\% & 87.65 \scriptsize{(90.6\%)} & 67.74 \scriptsize{(87.3\%)} & 58.71 \scriptsize{(95.9\%)} & 91.3\% \\ \midrule

DivPrune \scriptsize(CVPR'25) & 86.12 \scriptsize{(92.2\%)} & 71.36 \scriptsize{(87.5\%)} & 58.07 \scriptsize{(93.9\%)} & 91.2\% & 89.98 \scriptsize{(93.1\%)} & 61.88 \scriptsize{(79.8\%)} & 58.06 \scriptsize{(94.8\%)} & 89.2\% \\
w/ \proposed{} & 91.36 \scriptsize{(97.8\%)} & 74.11 \scriptsize{(90.9\%)} & 60.42 \scriptsize{(97.7\%)}& 95.5\% & 90.75 \scriptsize{(93.8\%)} & 73.75 \scriptsize{(95.1\%)} & 59.13 \scriptsize{(96.6\%)} & 95.2\% \\ \midrule

VisionZip \scriptsize(CVPR'25) & 90.41 \scriptsize{(96.8\%)} & 77.10 \scriptsize{(94.5\%)} & 60.68 \scriptsize{(98.2\%)} & 96.5\% & 88.79 \scriptsize{(91.8\%)} & 61.21 \scriptsize{(78.9\%)} & 58.36 \scriptsize{(95.3\%)} & 88.7\% \\
w/ \proposed{} & 92.08 \scriptsize{(98.6\%)} & 77.20 \scriptsize{(94.6\%)} & 61.27 \scriptsize{(99.1\%)} & 97.4\% & 90.19 \scriptsize{(93.3\%)} & 66.29 \scriptsize{(85.4\%)} & 59.20 \scriptsize{(96.7\%)} & 91.8\% \\ \specialrule{0.1em}{0em}{0em}

%% Retain 22.2%
\rowcolor{gray!20} \multicolumn{9}{c}{\textit{Retain 22.2\% Tokens ($\downarrow$ 77.8\%)}} \\ 
FastV \scriptsize(ECCV'24) & 83.69 \scriptsize{(89.6\%)} & 72.35 \scriptsize{(88.7\%)} & 57.09 \scriptsize{(92.3\%)} & 90.2\% & 80.35 \scriptsize{(83.1\%)} & 63.18 \scriptsize{(81.4\%)} & 54.84 \scriptsize{(89.5\%)} & 84.7\% \\
w/ \proposed{} & 91.27 \scriptsize{(97.7\%)} & 74.47 \scriptsize{(91.3\%)} & 60.29 \scriptsize{(97.5\%)} & 95.5\% & 85.74 \scriptsize{(88.7\%)} & 66.03 \scriptsize{(85.1\%)} & 56.30 \scriptsize{(91.9\%)} & 88.6\% \\ \midrule

DivPrune \scriptsize(CVPR'25) & 83.02 \scriptsize{(88.9\%)} & 70.04 \scriptsize{(85.9\%)} & 56.47 \scriptsize{(91.3\%)}& 88.7\% & 86.50 \scriptsize{(89.5\%)} & 55.77 \scriptsize{(71.9\%)} & 56.04 \scriptsize{(91.5\%)} & 84.3\% \\
w/ \proposed{} & 90.79 \scriptsize{(97.2\%)} & 72.87 \scriptsize{(89.3\%)} & 59.81 \scriptsize{(96.7\%)} & 94.4\% & 87.19 \scriptsize{(90.2\%)} & 69.39 \scriptsize{(89.4\%)} & 57.09 \scriptsize{(93.2\%)} & 90.9\% \\ \midrule

VisionZip \scriptsize(CVPR'25) & 88.65 \scriptsize{(94.9\%)} & 71.81 \scriptsize{(88.0\%)} & 59.46 \scriptsize{(96.2\%)}& 93.0\% & 84.54 \scriptsize{(87.4\%)} & 51.04 \scriptsize{(65.8\%)} & 56.54 \scriptsize{(92.3\%)} & 81.8\% \\
w/ \proposed{} & 90.27 \scriptsize{(96.6\%)} & 71.97 \scriptsize{(88.2\%)} & 60.18 \scriptsize{(97.3\%)}& 94.0\% & 88.07 \scriptsize{(91.1\%)} & 62.06 \scriptsize{(80.0\%)} & 58.16 \scriptsize{(95.0\%)} & 88.7\% \\\specialrule{0.1em}{0em}{0em}

\end{tabular}
}
\vspace{-3ex}
\end{table*}

\vspace{-1ex}
\section{TFLOPs Analysis and Calculation}
\label{app:K_TFLOPs}
\vspace{-1ex}
We evaluate the computational efficiency of \proposed{} using TFLOPs, which quantifies the total floating-point operations during inference, following~\cite{FastV, SparseVLM, sparsevila, DivPrune}. Using Qwen3-VL-4B~\cite{qwen3vl} as the backbone ($L = 36$ layers, hidden dimension $d = 2560$, intermediate dimension $m = 9728$), we calculate TFLOPs for the prefill and decoding stages as follows:

\begin{equation}
\begin{aligned}
\text{FLOPs}_{\text{prefill}} &= L \times (4nd^2 + 2n^2d + 2ndm) \\
\text{FLOPs}_{\text{decoding}} &= \sum_{i=0}^{O-1} L \times (4d^2 + 2(n+i)d + 2dm)
\end{aligned}
\end{equation}
where $n$ is the number of input tokens and $O$ is the number of generated tokens. We assume a generation length of 1,000 tokens for this analysis. 

\begin{wraptable}{r}{0.52\textwidth}
\centering
\vspace{-25pt}
\caption{TFLOPs comparison on Qwen3-VL-4B.}
\label{tab:tflops_qwen3}
\small
\setlength{\tabcolsep}{5pt}
\resizebox{\linewidth}{!}{
\begin{tabular}{ccccc}
\toprule
\multirow{2}{*}{Ratio} & \multirow{2}{*}{Method} & Prefill & Decoding & Total \\
 &  & (TFLOPs) & (TFLOPs) & (TFLOPs) \\ \midrule
100\% & Vanilla & 1.637 & 2.158 & 3.795 \\ \midrule
\multirow{2}{*}{33.3\%} & FastV & 0.546 & 2.130 & 2.676 \\
 & w/ \proposed{} & 0.546 & 2.137 & 2.683 \\ \midrule
\multirow{2}{*}{22.2\%} & FastV & 0.363 & 2.126 & 2.489 \\
 & w/ \proposed{} & 0.363 & 2.133 & 2.496 \\ \bottomrule
\end{tabular}
}
\vspace{-25pt}
\end{wraptable}
\noindent
Since \proposed{} operates as a plug-and-play module during decoding, its prefill-stage cost is identical to that of the baseline pruning method.
As shown in~\cref{tab:tflops_qwen3}, the additional TFLOPs introduced by \proposed{} over the static FastV baseline is negligible at both retention ratios.
This confirms that \proposed{} preserves the efficiency benefits of token pruning while enabling adaptive visual token retrieval during decoding.

\vspace{-1ex}
\section{Impact of Decoding Token Budget}
\label{app:L_Decoding_token_budget}
\vspace{-1ex}

\begin{wrapfigure}{r}{0.45\textwidth}
  \centering
  \vspace{-15pt}
  \includegraphics[width=1\linewidth]{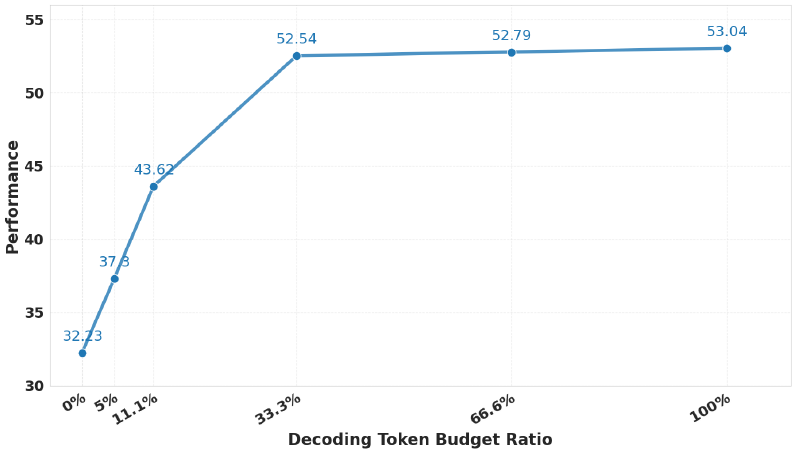}
  \vspace{-3.5ex}
  \caption{{Performance of \proposed{} on MathVerse across various decoding token budgets, with FastV as the pruning baseline.}}
  \label{fig:addon}
  \vspace{-15pt}
\end{wrapfigure}
In the CPTS module, the number of newly re-selected tokens $X^{\text{Pruned}, l}_v$ during decoding is set to $k$ by default, matching the prefill-stage budget of $X^{\text{Pruned}}_v$. Here, we investigate whether this decoding budget can be further compressed. We fix the prefill budget at 33.3\% and vary the size of $X^{\text{Pruned}, l}_v$ from 0\% to 100\% of the total tokens $N_v$, where 0\% corresponds to the vanilla FastV baseline without token swapping.
As shown in~\cref{fig:addon}, a decoding budget of 33.3\% already achieves performance comparable to the 100\% full-retention upper bound, indicating that a matched budget is sufficient for complex reasoning without the redundancy of the full visual context. Moreover, even under an extremely restricted budget of 11.1\%, \proposed{} maintains a substantial margin over the vanilla baseline.
This demonstrates that a matched decoding budget of 33.3 is sufficient for effective reasoning, while the strong performance at even lower budgets suggests the potential for further compression when prioritizing efficiency.

\vspace{-1ex}
\section{Detailed Related Works} \label{app:M_Detailed_Related_Works}
\vspace{-1ex}

\subsection{Efficient Multimodal Large Language Models}
Research for Efficient Multimodal Large Language Models (MLLMs) has emerged to address the significant challenges in inference speed and memory consumption caused by their massive scale. These studies can be broadly divided into two categories:
\textbf{1) Changing MLLM Internals}: These methods modify the architecture or the parameters of the model itself. 
Common techniques include 
\textbf{Quantization}~\cite{AWQ, OWQ, QLORA}—reduceing the precision of the numbers the model uses (e.g., moving from $16$-bit to $8$-bit or even $4$-bit) to save space;
\textbf{Distillation}~\cite{CompoDistill, LLaVA_KD, Vlsl, GenRecal, Masters, RIL}—where a large teacher model helps a smaller student model learn to perform just as well; and \textbf{Layer Pruning}~\cite{LaCo, ShortLLaMA, MKA}—identifying and removing parts of the neural network that do not contribute significantly to the final result.
\textbf{2) Optimizing the Inference Stage:} These methods focus on how a model can be used efficiently without necessarily changing its underlying weights. Key strategies include
\textbf{Token Pruning}~\cite{FastV, DivPrune, visionselector, VisionZip, VisionTrim, PDrop}—removing redundant or unnecessary visual tokens to reduce the workload for the model.
\textbf{Speculative Decoding}~\cite{SelfJudge, SpeculativeDecoding, Eagle, Medusa}—using a smaller, faster model to draft the next few tokens, which the larger model then quickly verifies to speed up the generation process.

\subsection{Visual Token Pruning for MLLMs}

\textbf{Methods.} 
To overcome the computational bottlenecks of MLLMs, visual token pruning has developed into two primary categories. \textbf{1) Training-based Methods:} These techniques, such as LLaVolta~\cite{llavolta}, ZipR1~\cite{ZipR1}, and VCM~\cite{vcm}, add learnable components to the model to help it reduce data, but they often require significant extra time and power to train. 
\textbf{2) Training-free Methods:} These are more widely used because they work instantly on existing models without additional training. 
These methods generally use three strategies. 
\textbf{Vision-Encoder Based}—using features directly from the vision-processing part of the model, like LLaVA-PruMerge~\cite{llava-prumerge}, DivPrune~\cite{DivPrune} and VisionZip~\cite{VisionZip}; 
\textbf{LLM-Based}—identifying important tokens using the internal attention mechanism of the language model, such as FastV~\cite{FastV}, ZipVL~\cite{ZipVL}, PDrop~\cite{PDrop}; and
\textbf{Cross-Modal}—looking at how the text query and image interact to decide what to keep, as seen in SparseVLM~\cite{SparseVLM} and SparseVILA~\cite{sparsevila}.

\smallskip \noindent \textbf{Evaluation.} 
Recent benchmarks like UniPruneBench~\cite{VTP_various_task_2} and LLMC+~\cite{VTP_various_task_3} have shown that while pruning makes models faster, being too aggressive can lead to a severe loss of performance on tasks that require attention to small details. Most existing methods focus on speed but often accidentally throw away the critical visual information needed for a truly deep understanding of an image. Our research specifically focuses on complex visual reasoning, aiming to keep and retrieve this important information that is usually lost in static pruning. This allows the model to perform difficult logical tasks accurately without losing the benefits of being fast and efficient.

\subsection{Visual Reasoning in MLLMs}
As MLLMs advance, the focus of evaluation has shifted from simple questions—what is in this image?—to complex, step-by-step logical thinking. Unlike basic visual understanding~\cite{GQA, TextVQA, ScienceQA}, complex visual reasoning~\cite{mathverse, mathvision, wemath, dynamath,mmmu-pro,logicvista} requires the model to interpret subtle visual cues throughout a long sequence of logical steps. To measure this ability, several difficult benchmarks have been developed. 
\textbf{1) Visual Math Reasoning}—Benchmarks such as MathVerse~\cite{mathverse}, WeMath~\cite{wemath}, and DynaMath~\cite{dynamath} challenge models to solve mathematical problems by directly interpreting diagrams and charts rather than just relying on text shortcuts. Logical Reasoning~\cite{mmmu-pro, logicvista}—Tasks like LogicVista~\cite{logicvista} and MMMU-Pro~\cite{mmmu-pro} evaluate expert-level analytical capabilities, complex puzzles, and spatial navigation. These studies collectively highlight that successful complex reasoning requires precise and uncompromised visual details. Our work is motivated by the observation that while standard pruning often accidentally throws away the very information needed for these benchmarks, our \proposed{} framework is designed to rescue and retrieve those essential visual cues exactly when the model needs them during the reasoning process.

\vspace{-1ex}
\section{Limitations and Future Works}
\vspace{-1ex}
A potential limitation of \proposed{} is that the selected visual tokens are applied uniformly across all transformer layers, without accounting for layer-wise visual requirements~\cite{aircache_laterncy_ref1, DevilsInMiddleLayers}. 
Since early layers tend to process low-level features while deeper layers handle higher-order reasoning, tokens discarded during the prefill stage may still be needed by specific layers to perform advanced logical derivations.
Additionally, \proposed{} relies on fixed values for the detection threshold $\tau$ and context-preserving duration $L$. Although these constant parameters prove effective across our experiments, adapting them dynamically to the difficulty of each query could further improve the balance between performance and efficiency. As future work, we plan to explore Reinforcement Learning (RL)-based strategies that enable the model to autonomously determine when to trigger token swapping and how many tokens to retrieve, allowing for more fine-grained adaptation without incurring expensive training costs.

\vspace{-1ex}
\section{Qualitative Examples}
\vspace{-1ex}
This section presents qualitative comparisons between \proposed{} and FastV~\cite{FastV}, illustrating how adaptive token swapping during decoding helps the model capture the precise visual features needed at each reasoning step.

\begin{figure}[h]
    \centering
    \includegraphics[width=0.92\linewidth]{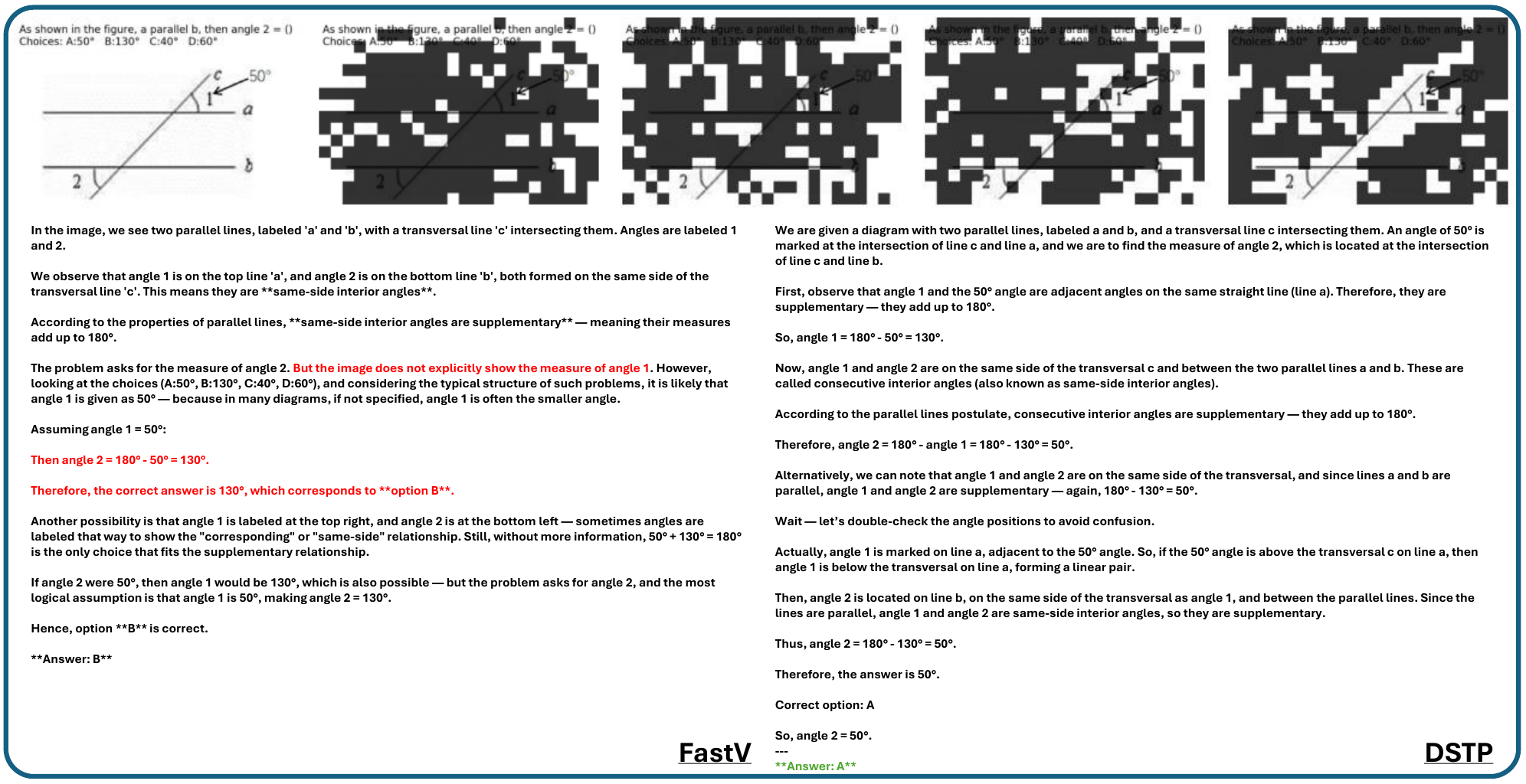}
    \vspace{-1ex}
    \caption{\footnotesize Example 1: Robustness in numeric information extraction. Comparison highlighting \proposed{}'s ability to maintain high fidelity in reading numerical figures where baseline models like FastV exhibit semantic drift.}
    \vspace{-5ex}
    \label{fig:app_example_1}
\end{figure}

\begin{figure}[h]
    \centering
    \includegraphics[width=0.92\linewidth]{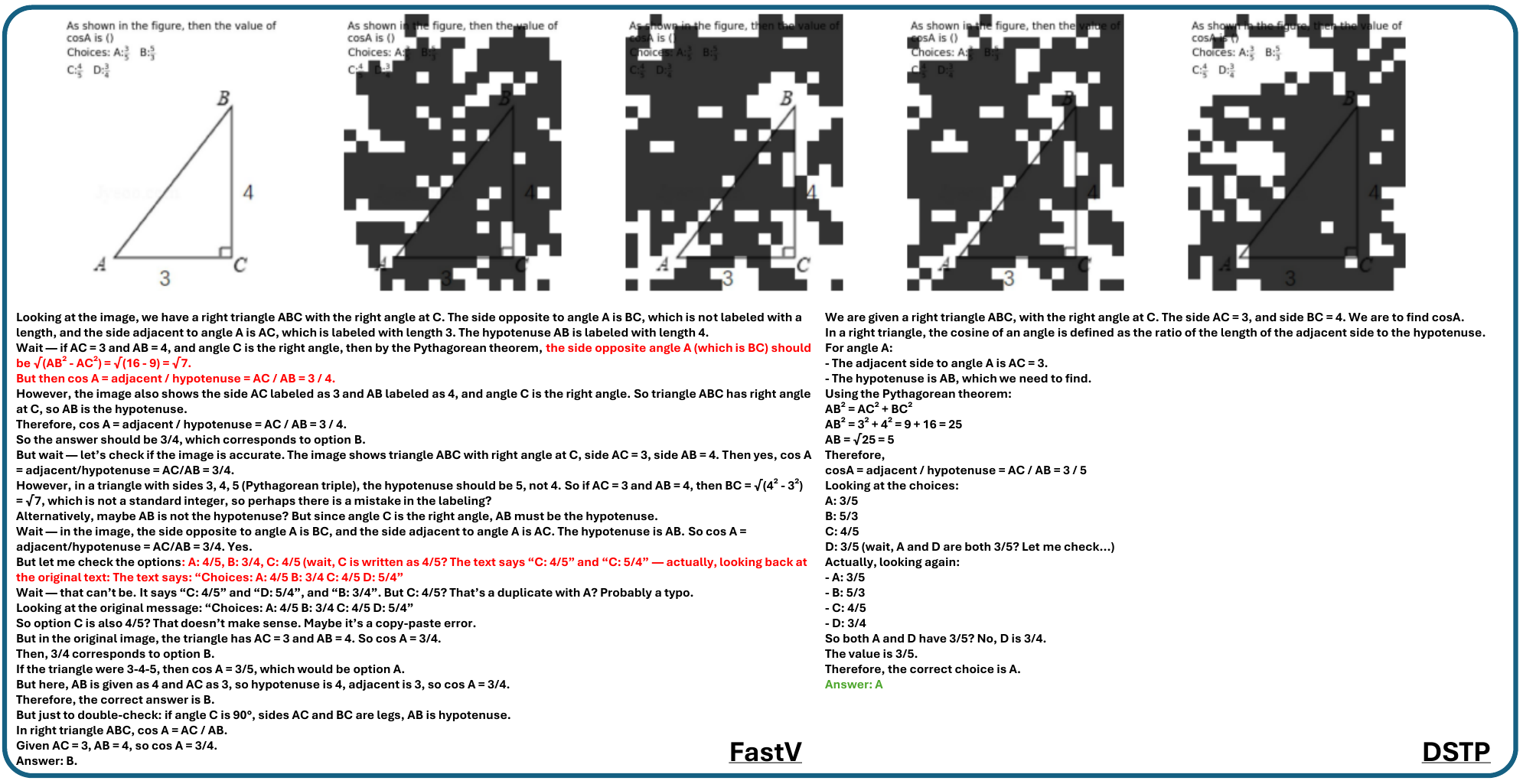}
    \vspace{-1ex}
    \caption{\footnotesize Example 2: Structural parsing of geometric primitives. \proposed{} demonstrates a deeper understanding of spatial layouts, correctly identifying structure, unlike FastV which misinterprets the geometric arrangement.}
    \vspace{-5ex}
    \label{fig:app_example_2}
\end{figure}

\begin{figure}[h]
    \centering
    % \vspace{-20ex}
    \includegraphics[width=0.92\linewidth]{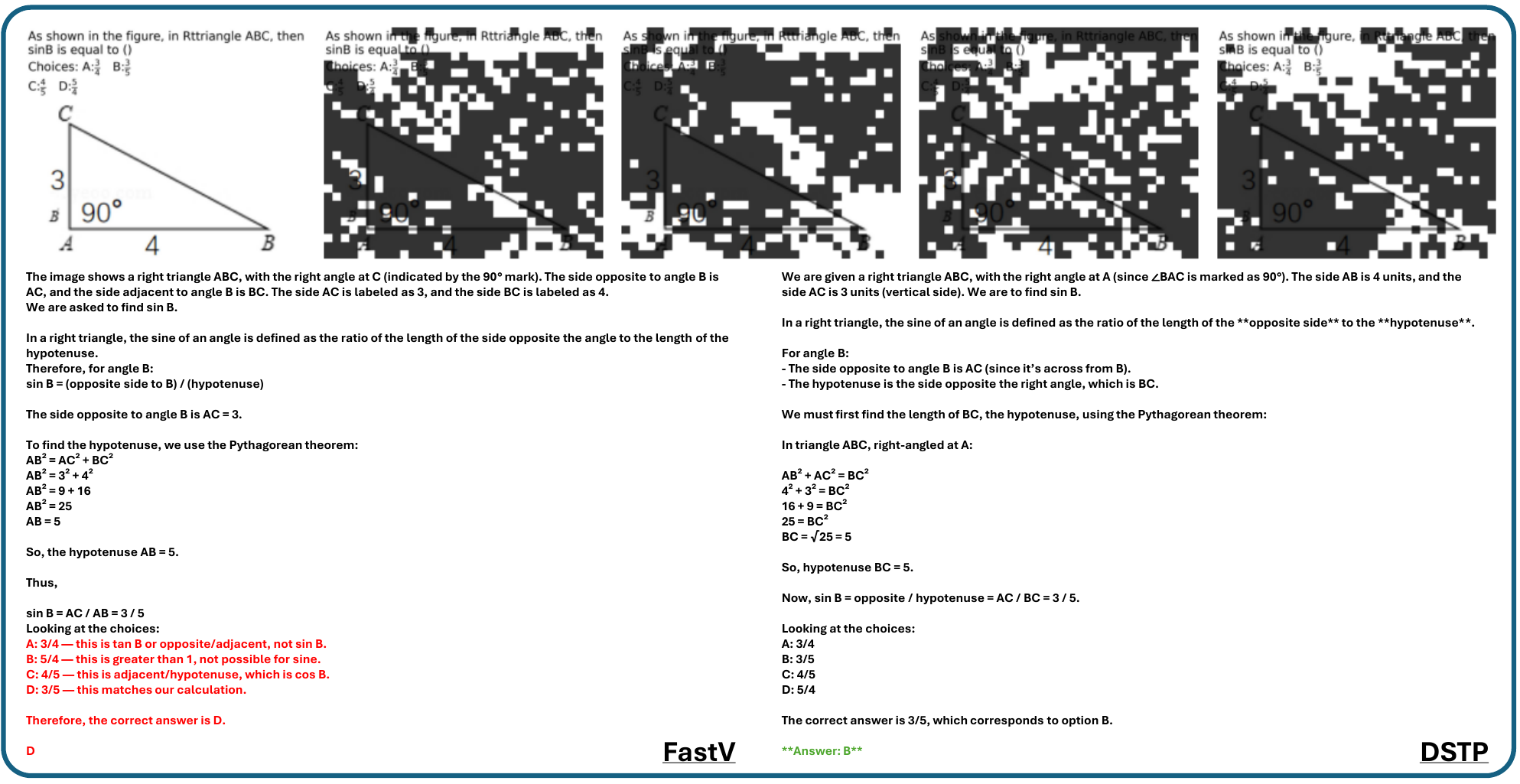}
    \vspace{-1ex}
    \caption{\footnotesize Example 3: Discerning selection states. Comparison of model performance on multiple-choice tasks. \proposed{} correctly recognizes the intended choice, while FastV fails to correlate the visual marker with the corresponding text.}
    \vspace{-5ex}
    \label{fig:app_example_3}
\end{figure}

\cref{fig:app_example_1} shows a case involving dense numeric information. FastV suffers from digit omission and semantic drift, whereas \proposed{} maintains high fidelity by dynamically retrieving the relevant visual tokens as the reasoning progresses.
\cref{fig:app_example_2} highlights geometric reasoning. FastV misidentifies the properties of a right-angled triangle, leading to flawed derivations, while \proposed{} correctly parses the geometric structure to produce sound inference.
Finally, \cref{fig:app_example_3} presents a failure mode where FastV solves the underlying mathematical problem correctly but fails to map the result to the correct answer option due to inaccurate perception of the choice labels. 
In contrast, \proposed{} retrieves the visual tokens corresponding to the answer choices at the final selection step, enabling the model to correctly identify and select the correct option.
Additionally, we provide qualitative examples from MMMU-Pro~\cite{mmmu-pro} to demonstrate the generalizability of our findings across diverse visual reasoning tasks.

\begin{figure}[h]
    \centering
    % \vspace{-20ex}
    \includegraphics[width=0.92\linewidth]{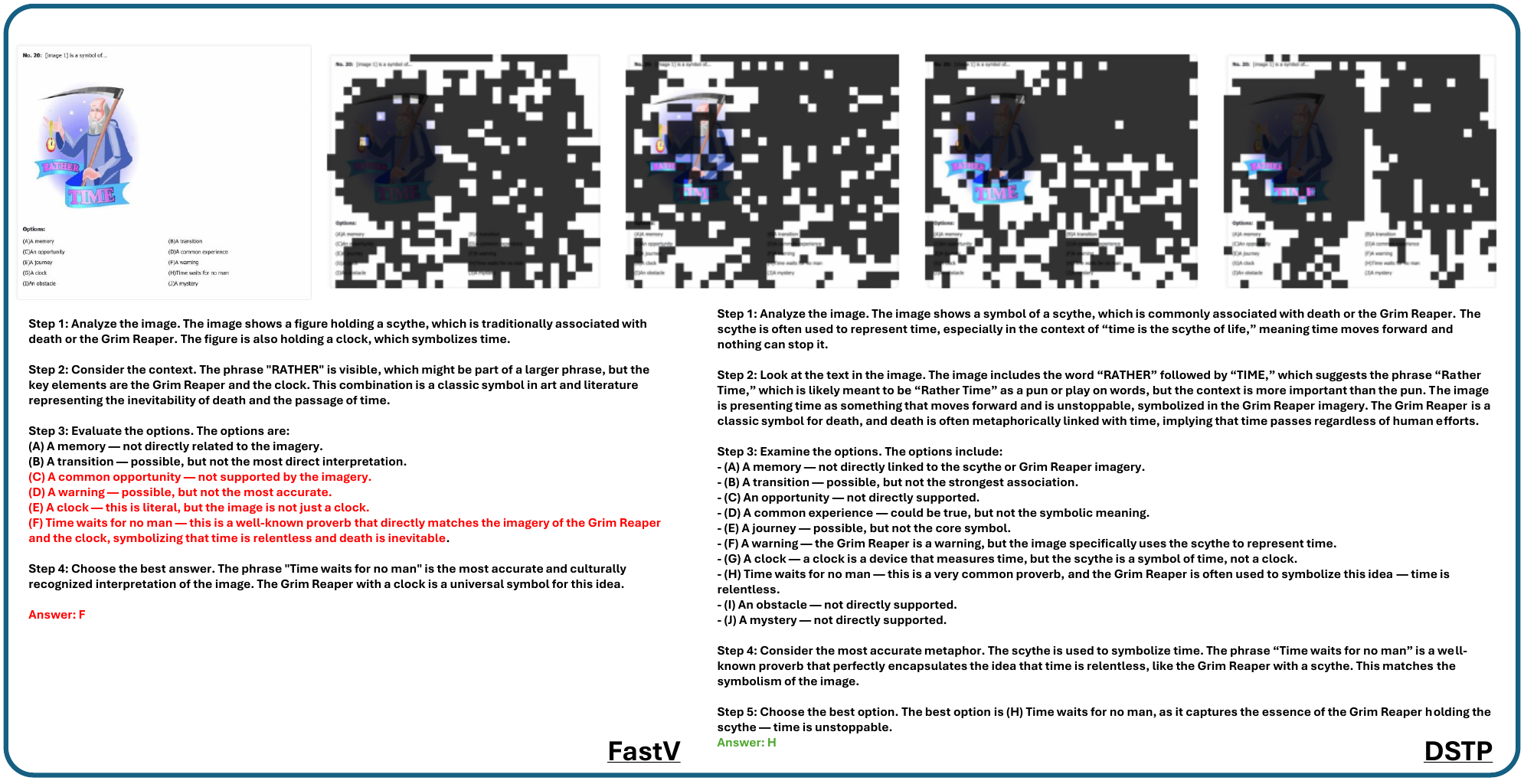}
    \vspace{-1ex}
    \caption{\footnotesize Example 4: Another discerning selection states.}
    \vspace{-3ex}
    \label{fig:app_example_4}
\end{figure}

\begin{figure}[h]
    \centering
    % \vspace{-20ex}
    \includegraphics[width=0.92\linewidth]{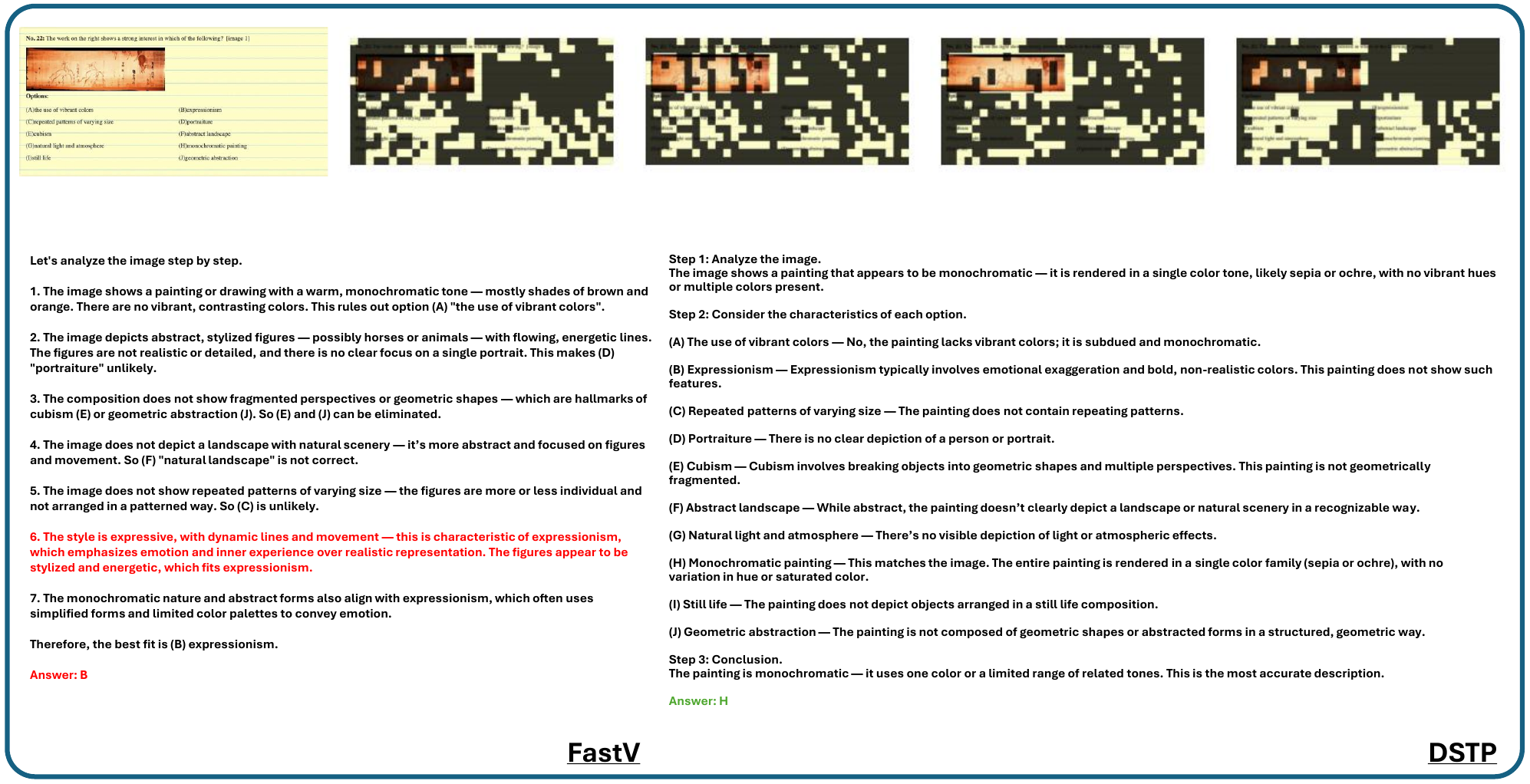}
    \vspace{-1ex}
    \caption{\footnotesize Example 5: Visual content recognition under token pruning. \proposed{} recognizes the painting itself, while FastV misunderstands the painting as related tokens are discarded.}
    \vspace{-5ex}
    \label{fig:app_example_5}
\end{figure}

\end{document}